\documentclass[fleqn,10pt]{wlscirep}
\usepackage[utf8]{inputenc}
\usepackage[T1]{fontenc}

\usepackage{lineno}

\usepackage{times}
\usepackage{helvet}
\usepackage{courier}

\usepackage{amsmath,amssymb,amsfonts}
\usepackage{algorithmic}
\usepackage{graphicx}
\usepackage{textcomp}
\usepackage{xcolor}
\usepackage{caption}
\usepackage{subcaption}
\graphicspath{ {./Figures/} }
\usepackage{tabularx}
\usepackage{hyperref}
\usepackage{algorithm}
\usepackage{algorithmic}
\usepackage{multirow}

\usepackage{cite}

\usepackage{makecell}

\usepackage{xcolor}
\usepackage{soul}
\usepackage{colortbl}
\usepackage{comment}

\usepackage{tikz}
\usepackage{pgfplots}
\pgfplotsset{compat=1.17}

\title{A high resolution urban and rural settlement map of Africa using deep learning and satellite imagery}

\author[1,2,5*]{Mohammad Kakooei}
\author[3,5]{James Bailie}
\author[4,5]{Markus B. Pettersson}
\author[4,5,+]{Albin Söderberg}
\author[4,5,+]{Albin Becevic}
\author[1,4,5]{Adel Daoud}
\affil[1]{Institute for Analytical Sociology, Linköping University, Sweden}
\affil[2]{Geomatics, Department of Environmental and Life Sciences, Karlstad University, Sweden}
\affil[3]{Department of Statistics, Harvard University, Massachusetts, USA}
\affil[4]{Department of Computer Science and Engineering, Chalmers University of Technology, Gothenburg, Sweden}
\affil[5]{The AI and Global Development Lab (www.global-lab.ai)}

\affil[*]{Corresponding author: Mohammad Kakooei (mohammad.kakooei@liu.se)}

\affil[+]{These authors contributed equally to this work.}

\begin{abstract}
Accurate and consistent mapping of urban and rural areas is crucial for sustainable development, spatial planning, and policy design. It is particularly important in simulating the complex interactions between human activities and natural resources. Existing global urban-rural datasets such as such as GHSL-SMOD, GHS Degree of Urbanisation, and GRUMP are often spatially coarse, methodologically inconsistent, and poorly adapted to heterogeneous regions such as Africa, which limits their usefulness for policy and research. Their coarse grids and rule-based classification methods obscure small or informal settlements, and produce inconsistencies between countries. In this study, we develop a DeepLabV3-based deep learning framework that integrates multi-source data, including Landsat-8 imagery, VIIRS nighttime lights, ESRI Land Use Land Cover (LULC), and GHS-SMOD, to produce a 10~m resolution urban-rural map across the African continent from 2016 to 2022. The use of Landsat data also highlights the potential to extend this mapping approach historically, reaching back to the 1990s. The model employs semantic segmentation to capture fine-scale settlement morphology, and its outputs are validated using the Demographic and Health Surveys (DHS) dataset, which provides independent, survey-based urban-rural labels. The model achieves an overall accuracy of 65\% and a Kappa coefficient of 0.47 at the continental scale, outperforming existing global products such as SMOD. The resulting High-Resolution Urban-Rural (HUR) dataset provides an open and reproducible framework for mapping human settlements, supporting UN Sustainable Development Goal (SDG) 11---Sustainable Cities and Communities---and enabling more context-aware analyses of Africa's rapidly evolving settlement systems, which indirectly support other SDGs, such as SDG 1 (No Poverty), by distinguishing human settlement types. We release a continent-wide urban-rural dataset covering the period from 2016 to 2022, offering a new source for high-resolution settlement mapping in Africa.

\end{abstract}
\begin{document}

\flushbottom
\maketitle
\thispagestyle{empty}

\section*{Introduction}
Rural and urban areas have many interconnections, but they play different socio-economic and environmental roles worldwide. Urban systems rely on external inputs such as electricity, fuels, and labor, while rural systems depend more on natural resources and as such demonstrate greater environmental sustainability \cite{wandl2014beyond}. While many previous studies have highlighted broad disparities between rural and urban areas, such as differences in infrastructure, health, education, and economic opportunity, these analyses often remain static and descriptive, lacking consideration of how urban systems evolve, interact, and contribute to spatial transformations across time \cite{tacoli2003links,simon2008urban}. In reality, urban and rural areas exist along a continuum of spatial and functional interdependence. Urban expansion reshapes surrounding rural regions through land-use conversion, migration flows, and market integration, while rural areas supply resources, labor, and ecosystem services essential to urban growth \cite{seto2012global,brenner2015towards}.

Urban areas systematically differ from their rural counterparts on multiple dimensions. For example, population density in urban areas is higher than in rural areas. Urban areas have advanced infrastructure and diverse economies, while rural regions typically depend on agriculture and natural resources and have fewer services. In addition, health and education services are generally better in urban areas than in rural regions. Despite these differences, urban and rural areas are interdependent---urban regions provide economic opportunities and services, and rural areas provide food, labor, and ecosystem benefits \cite{xiao2017understanding,cattaneo2022economic,liu2024novel}. Furthermore, travel time from rural areas to urban centers is a critical factor. Proximity to urban centers significantly impacts employment opportunities, access to services, and quality of life for rural communities, illustrating their interdependence and enhancing market access \cite{wandl2014beyond}.

Beyond descriptive analyses, several analytical studies have examined how urban and rural areas are both connected and differentiated. For example, Tuia et al. \cite{tuia2009detection} and He et al. \cite{he2021analysis} applied clustering techniques to analyze socio-economic gradients within urban environments. Similarly, Kim et al. \cite{kim2018enhancing} and Yhee et al. \cite{yhee2021gis} used GIS-based multi-criteria decision analysis to evaluate infrastructure accessibility in metropolitan regions of South Korea. In rural contexts, other studies have focused on livelihood strategies and resilience indicators (e.g., Ellis \cite{ellis2000determinants}; Alobo \cite{alobo2015rural}), while Wigley et al. \cite{wigley2020measuring} and Banke et al. \cite{banke2018assessing} investigated rural service accessibility through mixed-methods approaches. Together, these works provide valuable analytical insights and demonstrate the importance of frameworks capable of accommodating both rural and urban complexities within a unified system.

Maps of the geographical boundaries between urban and rural areas are the backbone of any such unified systems. For example, these maps play a key role in understanding environmental and climate impacts, particularly air pollution disparities. Ozone exposure varies significantly between urban and rural regions, resulting in differential health risks. While urban areas bear the highest pollution-related mortality burden, rural areas provide critical ecosystem services that help mitigate air quality degradation. Malashock et. al. \cite{malashock2022global}  found that ozone-attributable mortality in urban areas increased from 35\% to 37\% between 2000 and 2019, driven by higher levels of exposure and increased population density. In contrast, rural areas had lower direct health impacts but served as vital buffers for air quality regulation. These studies highlight the need for accurate urban-rural maps to analyze these interactions and thereby support effective policy interventions.

However, many urban-rural maps are based on administrative boundaries which reflect legacy political decisions rather than present-day settlement patterns. In contrast, land cover classification and Urban Morphological Zones (UMZs) provide clear data-driven methods for delineating urban and rural areas. These methods focus on spatial features like built-up density instead of just population counts, increasing the informativeness of the resulting urban-rural maps. Yes, despite extensive global efforts to map human settlements, existing data-driven urban-rural maps remain constrained by several critical limitations. First, spatial resolution is typically coarse (around 1~km), as is the case for GHSL-SMOD  \cite{pesaresi2016operating}, the GHS Degree of Urbanisation \cite{pesaresi2024advances}, and GRUMP  \cite{CIESIN2011_GRUMPv1}, which prevents accurate delimitation of small settlements and informal urban areas that are prevalent across Africa. Second, methodological inconsistency across datasets and countries results in divergent definitions of what constitutes ``urban'' and ``rural.'' National definitions of urban and rural areas vary, which makes global comparisons complicated---what is considered ``urban'' in one country may be classified as ``rural'' in another \cite{delgado2022interaction}. Third, temporal adaptability is limited. Many existing maps are produced for discrete years or long time intervals, which restricts their usefulness for tracking rapid urbanization dynamics or rural transformations. Finally, rule-based and population-driven approaches dominate prior studies, making them sensitive to census inaccuracies and incapable of capturing the visual or contextual nuances observable in satellite imagery. These gaps collectively underscore the need for a high-resolution, context-aware mapping framework which is based on spatial features and is capable of providing geographically consistent and temporally flexible urban-rural classifications. The present study addresses this need through a deep learning-based semantic segmentation approach that integrates multi-source satellite data to produce a unified 10~m urban-rural map for the African continent.

The potential benefits of the present study are multiple. High-resolution urban-rural maps help monitor urban expansion, analyze land-use changes and guide infrastructure and service planning. Accurate mapping of urban and rural areas is essential for sustainable development, resource management, and effective policymaking. For example, connectivity between urban and rural areas, especially through small and intermediate cities, provide significant benefits to rural populations. Cattaneo et. al. \cite{cattaneo2021global} emphasize that policymakers should prioritize investing in small and intermediate cities to improve access to services, which is especially important in the Global South, where 64\% of people reside in or near small cities. This conclusion demonstrates the importance of having high-resolution urban-rural maps---such as the one developed in the current paper---in regions like Africa.

Many African countries, particularly in Eastern and Central Africa, are in the early stages of rural economic development. Strengthening these economies could help retain labor and reduce pressures associated with urban migration. Effective migration management between rural and urban areas is a critical challenge for policymakers. For this, they require accurate urban-rural mapping, so that they can take into account regional differences in migration drivers instead of implementing less effective, uniform strategies \cite{christiaensen2022rural}. 

In addition to high-resolution classifications, the present work also provides a time-series mapping of urban-rural areas.
Such temporal mapping is essential for analyzing urbanization trends and addressing related challenges. For example, peri-urban areas---which are located between cities and the countryside---face challenges from rapid expansion, competing land uses, and inadequate infrastructure. In fast growing regions like Sub-Saharan Africa, these areas evolve quickly and require their own peri-urban governance reforms \cite{hutchings2022understanding}, emphasizing the need for time-series urban-rural maps that allow researchers to analyze areas as they transition from rural to urban.

In summary, the relationship between urban and rural areas is complex, involving resource interdependencies, socio-economic gaps, and environmental dynamics. Therefore, there is a critical need for high-resolution, time-series urban-rural mapping to support evidence-based policymaking. Current challenges such as inconsistent national classifications and official statistics; small and informal settlements; rapid evolution of peri-urban areas; and uneven migration show the limitations of existing products for mapping the African continent. The map proposed in the current paper addresses these challenges and assists researchers and policymakers to: monitor urban growth and land-use changes more accurately; detect transitional zones like peri-urban areas where infrastructure develops; evaluate the environmental impact of urban growth on rural sustainability; and tailor policies to regional needs, such as the support of small cities or the strengthening of agricultural hubs. By tracking spatial and temporal changes, this map can allow policies to be more responsive to the evolving urban-rural landscape.

\subsection*{Application to the Demographic and Health Surveys} 

Urban and rural classifications are included in the Demographic and Health Surveys (DHS) dataset. This enables researchers to do comparative analyses of development trends across these areas. For instance, a time-series analysis of the wealth index in Nigeria across DHS survey years, 1999, 2003, and 2008, shows values of 0.321, 0.438, and 0.466 in urban areas, while rural areas reported lower values of 0.162, 0.248, and 0.235, respectively \cite{oyekale2013assessment}. Another common application is the study of child mortality, which is consistently higher in rural areas compared to urban settings in low- and middle-income countries, highlighting the potentially beneficial effects of urbanization \cite{li2023association}. However, the utility of DHS data is constrained by its reliance on survey years and locations. Without spatially explicit urban-rural maps, these analyses cannot be extended to unsampled areas or periods between surveys. This highly limits their applicability for longitudinal or geographically comprehensive studies.

Having access to high-resolution, time-series rural-urban maps can significantly enhance the utility of existing survey datasets like DHS. For example, Pettersson et al. \cite{pettersson2023time} used DHS data to generate a poverty map for Africa. Their results showed that the deep learning model performed differently across rural and urban areas—achieving higher accuracy in rural regions and lower accuracy in urban ones. However, they did not incorporate rural-urban labels during inference or map generation, as such labels were not available consistently across time and space to be used as input features. Integrating rural-urban classifications at inference time would provide more informative features for machine learning models, potentially improving the prediction of wealth indices.

\subsection*{Available datasets}

The Global Human Settlement Layer (GHSL) project by the European Commission \cite{pesaresi2016operating} has produced a comprehensive dataset covering the period from 1975 to 2030 (in 5-year intervals) named GHSL-SMOD, which includes both historical data and future projections. This dataset, derived from satellite imagery and census data, offering maps at a 1~km resolution. It provides various metrics for settled areas (urban and rural) and experimental features like multiple-class land cover classification for more detailed analysis. However, GHSL-SMOD was not directly generated to address urban-rural classes. They combined GHSL built-up data and population grids (GHS-POP) to create this map \cite{freire2016development}. However, the GHS-POP map itself was derived by disaggregated census counts. Having the census population, it was distributed into grid cells proportionally to built-up area presence \cite{freire2016development}. Providing the global map with mnultiple epochs from 1975 is positive aspect from the time-series aspect. While this product offers valuable temporal depth and global coverage, its coarse spatial resolution severely limits applications that require fine-scale differentiation between urban, and rural settlements. Moreover, GHSL-SMOD was not originally designed to explicitly classify urban and rural areas, and it combines built-up and population grids to infer settlement types, which introduces definitional inconsistencies across regions.

There is another version of the data above, named ``GHS Degree of Urbanization Classification.'' This raster dataset represents a global, multi-temporal rural-urban classification, based on global gridded population and built-up surface data generated by the GHSL project for the epochs 1975-2030 in 5-year intervals. This product is an update of the data released in 2023 based on the updates of the GHS-BUILT-S and GHS-POP \cite{pesaresi2024advances} but inherits similar drawbacks, including low spatial resolution (1~km) and rule-based classifications that depend heavily on population density thresholds, rather than data-driven feature learning.

The Global Urban-Rural Mapping Project (GRUMP) provides urban and rural population grids which are derived from census and nightlights. This data set is produced by the Columbia University Center for International Earth Science Information Network (CIESIN) in collaboration with the International Food Policy Research Institute (IFPRI), the World Bank and the Centro Internacional de Agricultura Tropical (CIAT) \cite{CIESIN2011_GRUMPv1}. However, GRUMP is now outdated, and its 1~km spatial resolution fails to capture small or low-light settlements common in many African regions. Additionally, reliance on nightlight intensity tends to overestimate bright rural areas and underestimate unlit or informal settlements, further reducing accuracy.

In conclusion, although several global datasets such as GHSL-SMOD, GHS Degree of Urbanization, and GRUMP provide valuable long-term coverage of urban and rural patterns, they do not fully satisfy the requirements for precise studies. Their spatial resolution (typically 1km) is too coarse for detailed analysis, which limits the applications that require finer-scale insights. Moreover, the definition of urban and rural areas in these datasets is largely rule-based, relying on census, population grids, and nightlight data, rather than using advanced machine learning approaches. As a result, these datasets are insufficient for studies requiring accurate, data-driven differentiation between urban and rural areas.

\section*{Materials and methods}

Two main approaches, including shallow models and deep learning models, are commonly used for large-scale map generation and semantic segmentation tasks. Historically, prior to the availability of high-resolution satellite imagery and increased computational capacity, shallow models dominated remote sensing applications. For instance, Fourier analysis applied to Advanced Very High Resolution Radiometer data was used for land-cover classification in Brazil \cite{andres1994fourier}; random forests combined with MODIS observations enabled global land-cover mapping \cite{sulla2019hierarchical}; and Landsat or Sentinel imagery with Random Forests (RF) supported regional cropland mapping in Canada \cite{amani2020application}. Similarly, artificial neural networks have been used for generating EU land-cover maps from Sentinel data \cite{mirmazloumi2022elulc}. While these approaches laid foundational work in satellite-based land-cover mapping, shallow learners rely primarily on hand-crafted features, fixed spatial scales, and limited contextual information. As highlighted by Zhu et al. \cite{zhu2017deep}, such models struggle to capture the multi-scale, hierarchical, and spatially complex patterns inherent in human settlements and natural landscapes, especially in heterogeneous or rapidly changing regions. Their generalization capacity across diverse geographies also remains limited, often requiring region-specific tuning or feature engineering.

The availability of temporal satellite imagery on a global scale, coupled with advancements in computational power and ML techniques, has led to a shift from shallow learning to deep learning in remote sensing. Modern deep learning models, particularly convolutional neural networks (CNNs) and semantic segmentation architectures, learn rich spatial representations directly from raw imagery, which enables more accurate and scalable mapping across large and diverse regions \cite{kakooei2023spatial,daoud2023using}. The most prevalent deep learning approach for land cover classification employs encoder-decoder network architectures, such as the U-Net architecture \cite{ronneberger2015u}, which has gained prominence in semantic segmentation tasks \cite{wang2022uctransnet}. Various backbones have been developed and applied to land cover classification with promising results \cite{chantharaj2018semantic, badrinarayanan2017segnet, Deeplabv3+, LoopNet, PolSarInceptExceptSeg, hamida2017deep}. For instance, Garg et al. \cite{Deeplabv3+} demonstrated that the DeepLabV3+ network architecture significantly outperformed traditional shallow models across all benchmarks, even when using a small dataset.

In this work, we generated a three-class High-Resolution Urban-Rural (HUR) map of the African continent at 10~m spatial resolution, using deep learning models and satellite imagery. This map classifies areas into urban, rural, and non-human settlement (NonHS) categories, covering the period from 2016 to 2022. Figure~\ref{fig:flowchart} presents a schematic overview of the methodology used to generate the HUR map. To construct synthetic target variables, we combined the ESRI LULC map with the JRC SMOD map. As independent variables, we used yearly median composites from Landsat-8 along with nighttime light data. The use of nighttime lights is motivated by earlier rule-based approaches that relied on such data to generate coarse-resolution maps. Similarly, Landsat data have historically been used in settlement mapping. However, unlike rule-based methods, our approach directly inputs raw Landsat and nighttime light data into a deep model, allowing the model to automatically learn urban-rural distinctions. We trained a deep model with land cover maps including urban and rural classes as intermediate variables. A key contribution of this study is that the classification of urban and rural areas accounts for contextual relationships beyond traditional rule-based definitions. For example, a small settlement surrounded by cropland or vegetation is more likely to be classified as rural rather than urban. This context-aware approach enables finer and more accurate delineation compared to rule-based methods. Guided by this insight, we leveraged deep learning models to integrate LULC and settlement data, producing high-resolution urban-rural maps.

Finally, to minimize artificial boundaries between tiles, the HUR map was generated using a sliding-window approach. In the final product, all classes other than urban and rural categories were labeled as NonHS, resulting in a three-class map. For evaluation, we used an independent dataset with rural and urban labels from the Demographic and Health Survey (DHS), comparing the HUR map against the JRC SMOD map and the synthetic target map. Since the DHS dataset includes intentional spatial displacement for privacy protection, we relied on an imputed version for evaluation purposes. The deep learning-based classification showed superior agreement with the DHS labels, supporting our hypothesis that incorporating neighborhood information is crucial for accurate urban-rural delineation. 

\subsection*{Study region}
Africa was selected as the study area for several reasons. First, the continent is experiencing the fastest urban growth in the world, with its urban population projected to increase by nearly 950 million by 2050. Much of this expansion is concentrated in small and medium-sized towns \cite{heinrigs2020africapolis}. However, traditional census data often fails to capture informal settlements, underscoring the need for advanced geospatial mapping to more accurately delineate urban-rural boundaries. Second, there are pronounced inequalities between urban and rural areas in Africa, particularly in access to healthcare, education, and infrastructure \cite{samuel2021decomposing}. This highlights the importance of precise spatial data to inform equitable resource allocation. Third, rapid urbanization in Africa exacerbates the impacts of climate change, including flooding, heat stress, and land degradation \cite{butterfield2017inspiring}. High-resolution urban-rural maps can therefore play a critical role in supporting climate resilience strategies and guiding sustainable land-use policies.

\subsection*{Data preparation}
\begin{figure}[!ht]
    \centering
    \includegraphics[width = 0.50\textwidth]{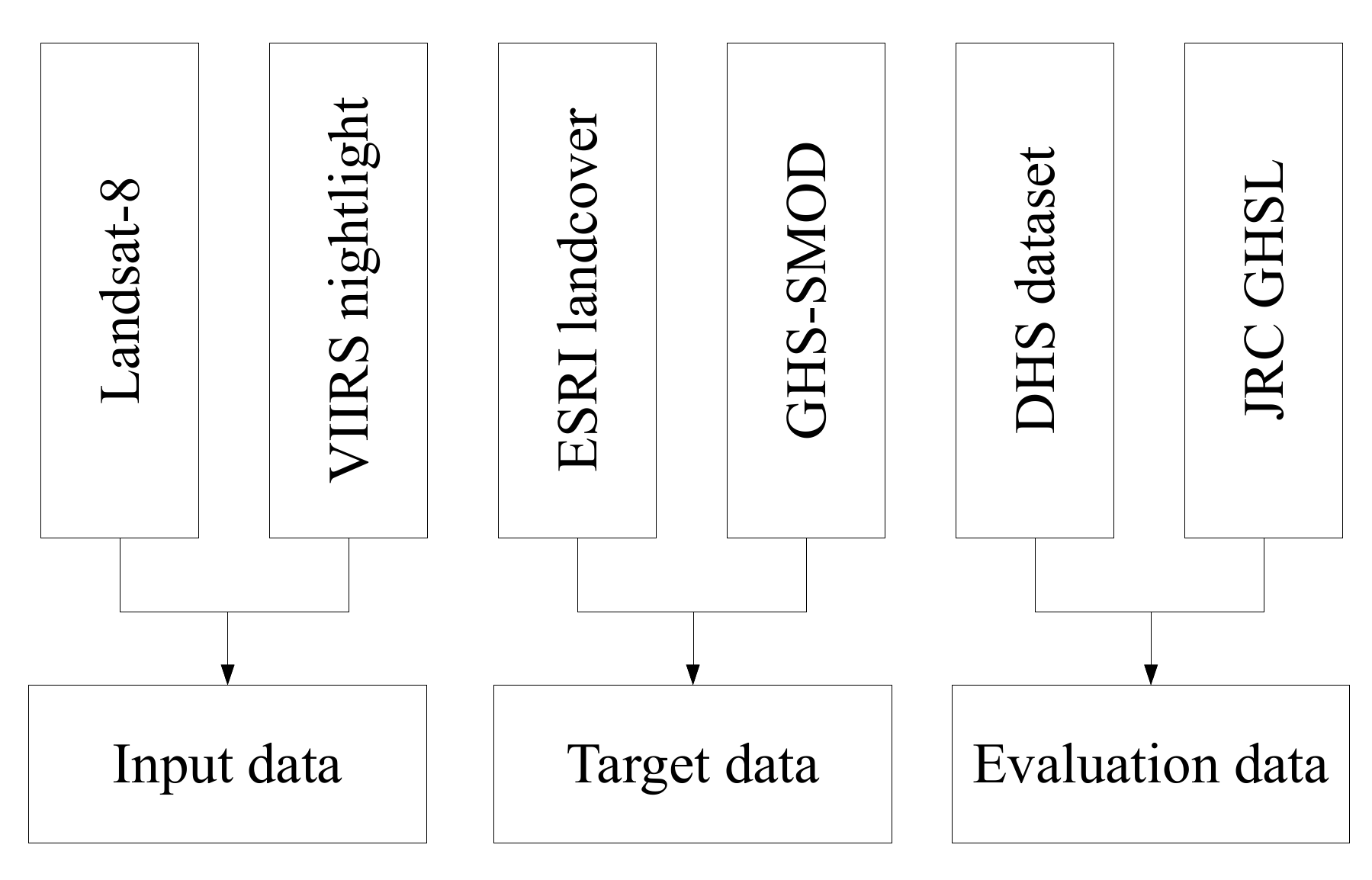}
    \caption{Data flows used to generate and evaluate our high-resolution urban-rural map.}
    \label{fig:dataflow}
\end{figure}

This study relies on remote sensing satellite imagery as its primary data source, accessed through Google Earth Engine (GEE) \cite{gorelick2017google}, a platform that facilitates the collection, storage, and analysis of large volumes of satellite data \cite{amani2020google}. Earth observation (EO) satellites collect data through regular and systematic orbits \cite{kansakar2016review}, but achieving a consistent number of high-quality observations over time is difficult due to issues such as cloud cover and shadows, which can degrade the imagery quality. Figure~\ref{fig:dataflow} illustrates the data flow in our research, outlining all the datasets used in this study.

\subsubsection*{Input data: Landsat and VIIRS nighttime light data}

The primary satellite imagery utilized in this study comes from Landsat-8. It provides multi-spectral data spanning from 2013 to the present. This makes it well-suited for our objective of generating an urban-rural map covering the years 2016 to 2022. Furthermore, as the Landsat mission program, initiated by NASA in 1974, has consistently deployed satellites to monitor and collect Earth surface data, allowing us to extend the temporal range of the dataset in the future. Landsat-8's core bands offer a resolution of 30~m $\times$ 30~m, capturing detailed information about the Earth's surface. In addition to these core bands, Landsat-8 includes a panchromatic band with 15~m $\times$ 15~m resolution and thermal bands with resolutions ranging 60--120~m \cite{LandsatGuide}. While Landsat-8 offers 15~m resolution in the panchromatic band, this study primarily relies on the 30~m and coarser bands. This approach supports potential future expansions in dataset temporality aligned with the Landsat mission program, given that older Landsat satellites lacked a panchromatic band.

The Visible Infrared Imaging Radiometer Suite (VIIRS) provides satellite imagery that captures nightlight intensity and has been available since 2012 with a resolution of 463~m \cite{hall2001history}. The Defense Meteorological Satellite Program (DMSP) also provides nighttime light data, covering the period from 1992 to 2013 at a coarser resolution of 927~m. VIIRS offers a superior spatial resolution compared to DMSP \cite{shi2014evaluating}, making it the only suitable choice for generating an urban-rural map for the years 2016 to 2022. Previous research has effectively utilized VIIRS data in combination with multi-spectral imagery to distinguish built-up areas (BuA) from other land cover types, such as forests, deserts, and vegetation, and to differentiate between rural and urban classes in Africa \cite{yeh2020using, pettersson2023time}. We anticipate that VIIRS nighttime light data enhance our model's capacity to differentiate urban and rural areas.

\subsubsection*{Target data: Land cover products and GHS-SMOD}
Land cover maps with synthetic generated urban-rural classes serve as intermediate variables. Accurate labeled target data is essential for incorporating these maps effectively. Utilizing pre-existing land cover maps is generally preferable to manually labeling a smaller custom dataset, which is a challenging task. Table~\ref{tab:landcover_datasets} provides details on existing land cover datasets, including their resolution and temporal coverage. This information was used to evaluate which datasets would be most suitable for serving as intermediate target labels in the mapping process.

\begin{table}[!ht]
\centering
\begin{tabular}{|l|l|l|l|}
\hline
  & \textbf{Dataset}        & \textbf{Resolution (m)} &\textbf{Temporality} \\ \hline
1 & ESA WorldCover            & 10 m      &   2020, 2021  \\
2 & ESRI Land Use/Land Cover & 10 m  &  2018--2022 \\
3 & NASA GlanCE-v001 & 30 m  & 2001--2019               \\
4 & GlobeLand30               & 30 m &   2000, 2010, 2020 \\
5 & From-GLC & 10/30 m      & 2010 \& 2015/2017          \\
6 & GlobCover & 300 m     & 2009           \\
\hline
\end{tabular}
\caption{An overview of the publicly-available land cover datasets with their resolution and temporal coverage.}
\label{tab:landcover_datasets}
\end{table}

The criteria for target data for this study was based on three properties: temporality, spatial coverage, and spatial resolution. Firstly, temporality: datasets with a long and consecutive temporal span allow for temporal evaluation, providing insights into the dataset's reliability over time. Secondly, spatial coverage: the dataset should ideally cover the entirety of Africa to capture the continent's diverse land surface features. Lastly, spatial resolution: higher spatial resolution datasets enable the generation of more detailed maps, which is crucial for precise urban-rural classification.

ESA WorldCover and ESRI Land Use Land Cover (LULC) stand out among land cover datasets due to their high spatial resolution of 10~m, which provides a detailed level of mapping. Of the two, ESRI LULC also offers superior temporal coverage.  From a temporal perspective, NASA's GlanCE-v001 \cite{arevalo2022global} provides extensive temporal coverage, its lower spatial resolution makes it unsuitable for this study's requirements. Similarly, GlobeLand30 \cite{chen2015global} has a good temporal span, but insufficient 30~m spatial resolution that falls short compared to the higher resolution needed. Consequently, this work proceeded with ESRI LULC due to its superior temporal coverage, higher estimated class accuracy, and better alignment with the deep learning approach, as it was produced using a deep model.

The ESRI LULC 2017--2022 Time Series dataset was selected as part of the target data to train the deep model. Developed by ESRI, a global leader in geographic information system (GIS) software, in partnership with Microsoft, this dataset provides a 10~m resolution global land cover time-series map covering the years 2017--2022. With an estimated accuracy of 85\%, this dataset was generated using a deep learning model for semantic segmentation, utilizing Sentinel-2 satellite data (10--60 meter resolution) as input. The dataset includes nine classes: Water, Trees, Flooded Vegetation, Crops, Built-up Area, Bare Ground, Snow/Ice, Clouds, and Rangeland.

GHS-SMOD is a component of the Global Human Settlement Layer (GHSL) project \cite{pesaresi2016operating}. It provides a human settlement map in a coarse resolution (1~km $\times$ 1~km) raster format, classifying grid cells by their degree of urbanization. This dataset is generated by combining various GHSL products, which include data on built-up surfaces and population estimates, and applying a clustering scheme detailed in Table~\ref{tab:smod_scheme}. Despite its coarse resolution, which limits its suitability for applications requiring medium to high-resolution human settlement data, GHS-SMOD can still serve as a valuable auxiliary dataset when merged with higher resolution labels. The dataset comprises eight classes: (1) Water Grid Cell; (2) Very Low Density Rural Grid Cell; (3) Low Density Rural Grid Cell; (4) Rural Cluster Grid Cell; (5) Suburban or Peri-urban Grid Cell; (6) Semi-Dense Urban Cluster Grid Cell; (7) Dense Urban Cluster Grid Cell; and (8) Urban Center Grid Cell.

\begin{table}[!ht]
    \centering
    \includegraphics[width=0.75\textwidth]{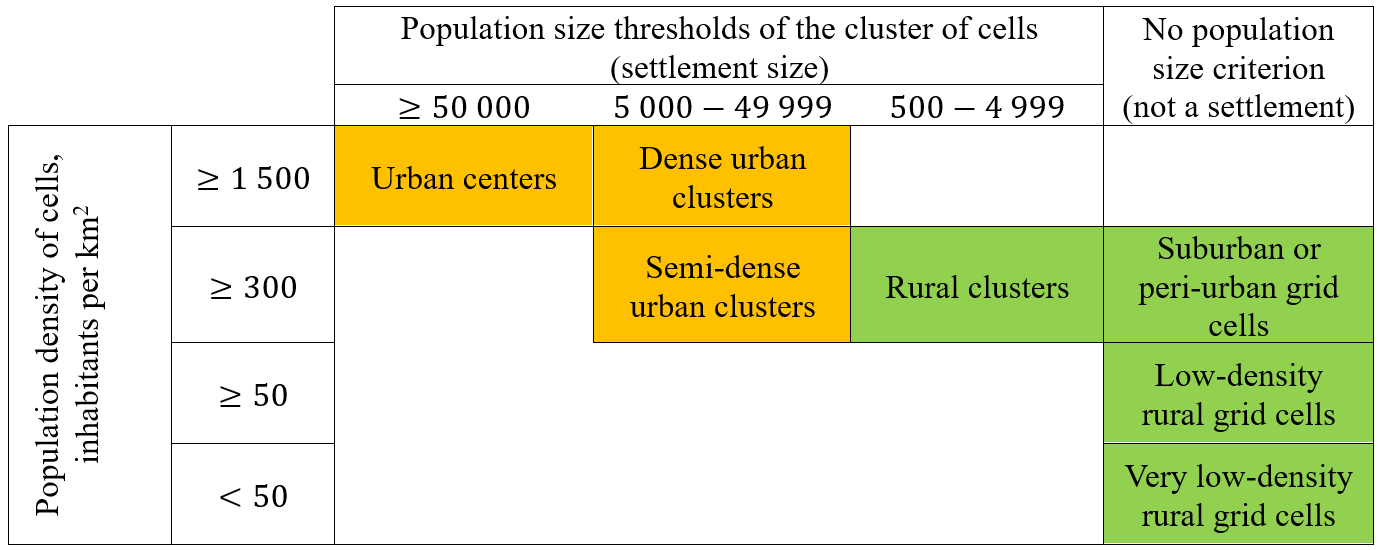}
    \caption{The classification scheme used for the GHS-SMOD product \cite{smod_scheme}. Which class a 1~km $\times$ 1~km cell is given is based both on the population density of the cell itself and the total population size of all neighboring cells of the same density category. According to this scheme, orange cells are considered as urban, while green cells are considered as rural.}
    \label{tab:smod_scheme}
\end{table}

\subsubsection*{Evaluation data: DHS dataset imputed by JRC HSL} \label{sec:DHS} 
The Demographic and Health Survey (DHS) is a global survey program conducted across multiple countries. In addition to collecting responses directly from participants, the dataset provides labels indicating whether survey locations are classified as rural or urban, along with displaced geographic coordinates. To protect privacy, DHS applies a spatial displacement of up to 2 km for urban points and 5 km for rural points before public release. To enable reliable evaluation, we used an imputed version of the DHS dataset. This imputation was derived using the JRC Human Settlement Layer (HSL) 2018 \cite{corbane2021convolutional}. The JRC developed GHS-S2Net, a deep learning framework designed to extract built-up areas from Sentinel-2 imagery at a 10~m resolution, employing a lightweight CNN architecture with a 5~px $\times$ 5~px image patch.

Assume the released DHS dataset $D^{DHS}$ contains points labeled as urban ($D^{Urban}$) and rural ($D^{Rural}$), such that

$$
D^{DHS} = \{D^{Urban}, D^{Rural}\}.
$$

Using the JRC HSL, each point is further assigned a label indicating whether it locates inside a human settlement (HS) or outside (NonHS). The dataset therefore becomes

$$
D^{DHS} = \{D^{Urban}_{HS}, D^{Urban}_{NonHS}, D^{Rural}_{HS}, D^{Rural}_{NonHS}\},
$$

where $D^{Urban}_{HS}$ shows urban-labeled points located within human settlement areas,  $D^{Urban}_{NonHS}$ shows urban-labeled points located outside human settlement areas, $D^{Rural}_{HS}$ shows rural-labeled points located within human settlement areas, and $D^{Rural}_{NonHS}$ shows rural-labeled points located outside human settlement areas.

For evaluation, we retain points with HS labels ($D^{Urban}_{HS}, D^{Rural}_{HS}$) as part of the urban and rural classes. Points with \(NonHS\) labels are treated as belonging to the non-human settlement (NonHS) class. To address the displacement issue in DHS data, an imputed version of the dataset was generated. Specifically, JRC HSL points within 2 km of released urban points and within 5 km of released rural points were selected to form $D^{Urban}_{NonHS,Imputed}$ from $D^{Urban}_{NonHS}$, and $D^{Rural}_{NonHS,Imputed}$ from $D^{Rural}_{NonHS}$.

These imputed datasets are then used as additional evaluation data for the urban and rural classes. Therefore, the summary of evaluation datasets are:

\begin{itemize}
    \item The urban dataset for the evaluation includes \(D^{Urban}_{HS}\) and \(D^{Urban}_{NonHS,Imputed}\).
    \item The rural dataset for the evaluation includes \(D^{Rural}_{HS}\) and \(D^{Rural}_{NonHS,Imputed}\).
    \item The non-human settlement dataset for the evaluation includes \(D^{Urban}_{NonHS}\) and \(D^{Rural}_{NonHS}\).
\end{itemize}

\subsubsection*{Retrieving data through GEE} 
Various bands from Landsat-8 form the core of the input data for this study. Additionally, nightlight data from VIIRS is included to enhance the accuracy of human settlement classification. GHS-SMOD, a human settlement data product, is used as an auxiliary dataset for label augmentation. The ESRI LULC land cover data product is appended as the final band, serving as a part of the target data. The complete setup of bands is detailed in Table~\ref{tab:band_setup}. Notably, bands 1-7 comprise the model input, while bands 8 and 9 serve as the target labels during the training phase.

\begin{table}[!ht]
\centering
\begin{tabular}{|l|l|l|l|}
\hline
& \textbf{Band} & \textbf{Type} & \textbf{Resolution} \\ \hline
1 & Blue & Landsat                & 30 m \\
2 & Green & Landsat              & 30 m \\
3 & Red & Landsat                 & 30 m \\
4 & Near Infrared (NIR) & Landsat        & 30 m \\
5 & Shortwave Infrared 1 (SWIR1) & Landsat & 30 m \\
6 & Shortwave Infrared 2 (SWIR2) & Landsat & 30 m  \\
7 & Nightlights  & VIIRS          & 464 m \\
8 & GHS-SMOD 2020 & Human Settlement Map & 1000 m\\
9 & ESRI LULC 2018-2020 & Land Cover Map & 10 m \\ \hline
\end{tabular}
\caption{A description of the data bands downloaded from GEE and used as input and target data in training our model.}
\label{tab:band_setup}
\end{table}

A major source of uncertainty arises from mixed pixels, where footprint contains both built-up and non-built-up elements. Mixed pixels are a challenge in coarse-resolution remote sensing \cite{ small2004landsat}. In many African settings, particularly in peri-urban fringes and small rural settlements, a 30 m Landsat pixel frequently includes roofs, bare soil, and vegetation within the same footprint. The resulting spectral signature becomes a composite of multiple materials, creating spectral mixing that blurs the distinction between settlement and non-settlement classes \cite{li2019mapping}. When such pixels are assigned a single binary label during training, they introduce label noise that the network cannot fully disentangle \cite{rolnick2017deep}. This ambiguity disproportionately affects heterogeneous transition zones. As noted in prior studies, classifiers tend to over-predict built-up areas where bright soil or metallic roofing materials dominate the pixel, while under-predicting sparse or vegetation-obscured settlements \cite{bhatti2014built}.

\begin{figure}[!ht]
    \centering
    \begin{subfigure}{.17\textwidth}
      \centering
      \includegraphics[width=.98\linewidth ]{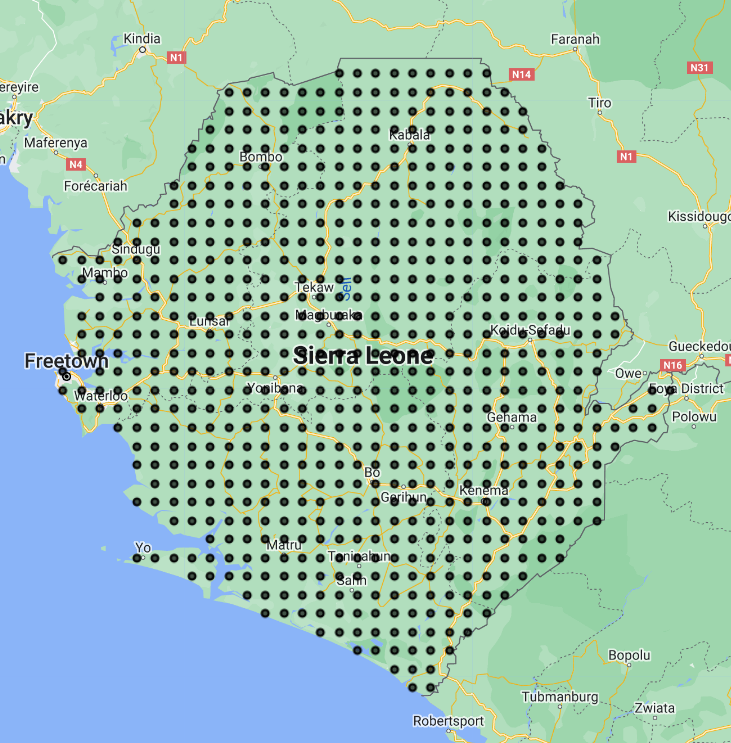}
      \caption{ }
    \end{subfigure}%
    \begin{subfigure}{.17\textwidth}
      \centering
      \includegraphics[width=.98\linewidth ]{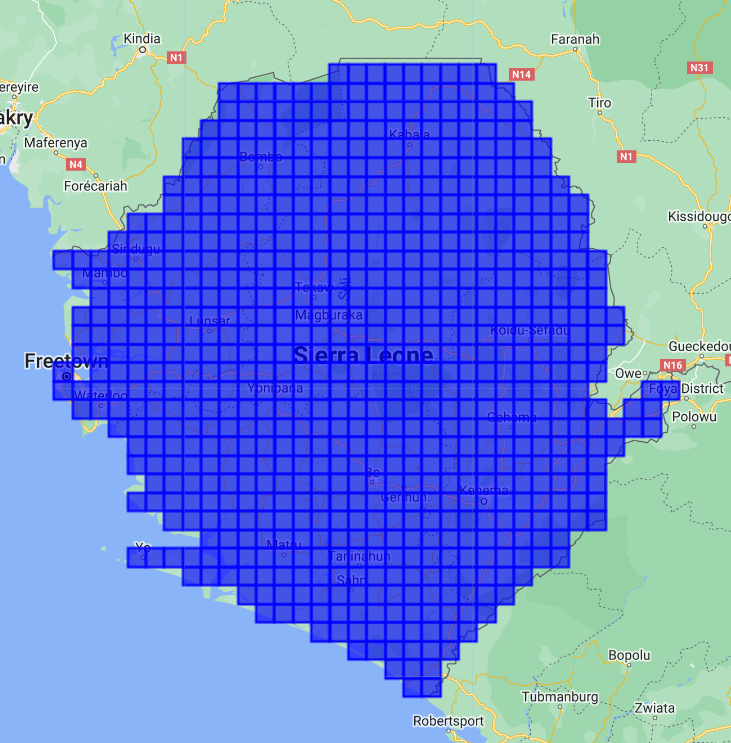}
      \caption{ }
    \end{subfigure}%
    \begin{subfigure}{.6\textwidth}
      \centering
      \includegraphics[width=.98\linewidth ]{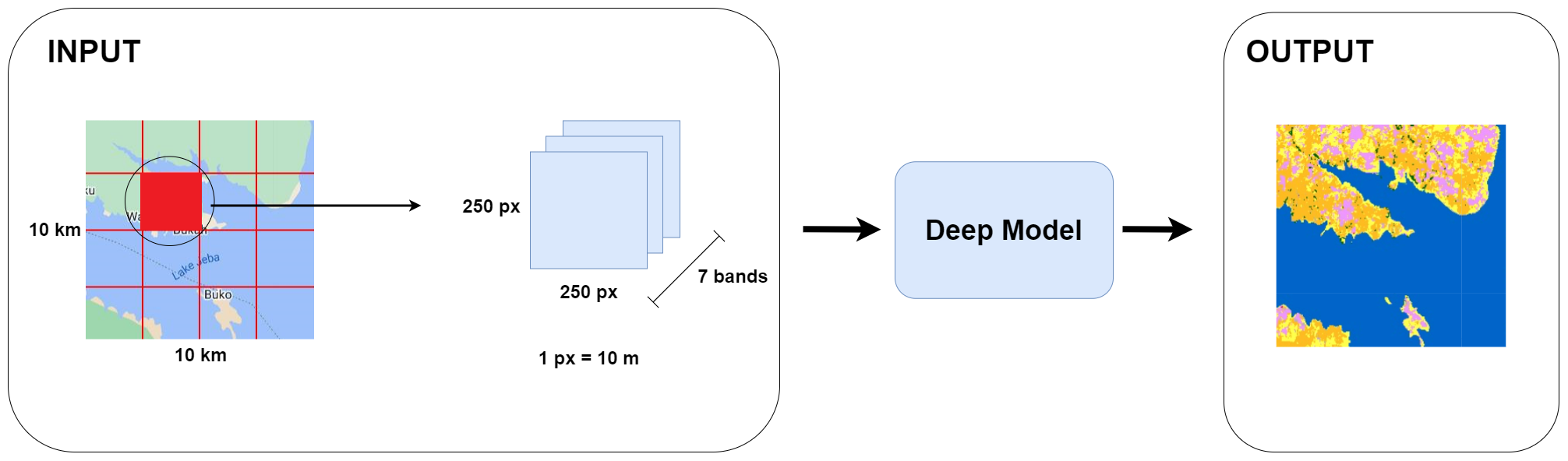}
      \caption{ }
    \end{subfigure}
    \caption{An illustration of how the HUR maps were generated, using Sierra Leone as an example: \textbf{(a)} the points from which images are sampled; \textbf{(b)} the image tiles covering the country; \textbf{(c)} a visualization of the format in which data is fed to the model.}
    \label{fig:data_gathering_model_input}
\end{figure}

Image data covering the entire surface of the continent of Africa was gathered, to form the training set. This to ensure a distribution of land cover classes that is as representative as possible for the continent as a whole. It also ensures that the model sees a sufficiently large set of land cover examples during training. In order to manage the organization of data, image tiles were retrieved and stored country-wise. The data retrieval process would follow the same procedure for every country. First, a number of points (as many as possible) were distributed across the country, spaced with a distance of 10 km, forming a grid. An example of this can be seen in Figure~\ref{fig:data_gathering_model_input}(a). Next, an image tile would be sampled from each point, in a 10x10 km bounding square. Together, the image tiles cover the surface of the country, as observed in Figure~\ref{fig:data_gathering_model_input}(b). Apart from the land cover labels (ESRI LULC) which represent an entire year, the satellite data itself is accessed as a collection of time-series images. A given location is revisited many times a year, meaning that the value of a pixel will vary between observations. An image composite method therefore had to be applied, to form a yearly image representation. For this purpose the best option was to take the median of all observed pixel values, as the mean can be affected by outliers. To ensure that the captured satellite imagery was of sufficient quality, a mask was used to filter away pixels covered by cloud or shadow. Ultimately, the dimensions of each retrieved tile was 10~km $\times$ 10~km or 1000~px $\times$ 1000~px (1~px = 10~m). Data are sampled at 10~m resolution, which is the native resolution of the target data. Images from all other bands (their native resolution shown in Table~\ref{tab:band_setup}) are resampled to 10~m resolution with nearest neighbor interpolation.

\subsection*{Research methods}

\subsubsection*{Semantic segmentation task}
Let $Y_{g}$ denote the true class label of a geo-location $g$, discretized into a pixel of size 10~m $\times$ 10~m in the satellite image $X_{g}$ captured at that location. The predicted class label for the same pixel, produced by the machine learning model $M$, is denoted as $\hat{Y}_{g}$. Thus, $\hat{Y}_{g} = M(X_{g})$ represents the label assigned by the model given the satellite imagery at point $g$. Representing the land surface as a raster of pixels is a practical simplification, but it comes with limitations. Each pixel is treated as homogeneous, while the actual land cover may be heterogeneous. In practice, a single pixel may encompass multiple classes, yet only one class label is assigned. As a result, the rasterized representation can oversimplify the true spatial complexity of land cover. This limitation underlines the contribution of this paper: generating a high-resolution urban-rural map at 10~m spatial resolution. Coarser datasets, such as GHS-SMOD, are provided at 1~km resolution, where each pixel represents a much larger area and thus suffers from greater oversimplification. By producing maps at finer resolution, we aim to reduce this loss of detail and improve classification accuracy.

\subsubsection*{Target data}
Generating an urban-rural map is a semantic segmentation task that requires pixel-wise labeled imagery. In this study, the models were trained on target data derived from ESRI LULC, supplemented with human settlement labels from GHS-SMOD. ESRI LULC provides high-resolution (10m) land cover classifications, while GHS-SMOD, despite its much coarser resolution (1~km), contributes large-scale differentiation between rural and urban settlement areas.

It is important to emphasize that both ESRI LULC and GHS-SMOD are outputs of deep convolutional neural networks and thus represent model predictions rather than direct ground truth observations. To address this limitation, the maps are evaluated against an independent ground truth dataset from DHS surveys described in the previous section. Consequently, the target data used for training in this study should be regarded as an intermediate variable rather than a definitive ground truth.

Label augmentation is a crucial and intricate part of preprocessing. ESRI LULC includes a `Built Area' class that aggregates all man-made structures such as buildings, roads, and other artificial features. While this general classification suffices for many land cover applications, it lacks the granularity needed for detailed human settlement analysis. To address this, a more refined classification scheme is necessary. For this purpose, a preprocessing method was devised to split the `Built Area' class into more informative sub-classes. This method utilizes the auxiliary dataset GHS-SMOD (2020), which classifies land surface by degree of urbanization. By overlaying the ESRI LULC labels with those from GHS-SMOD, all pixels classified as `Built Area' are relabeled according to the corresponding GHS-SMOD class at that location. GHS-SMOD provides seven specific settlement classes, but these are not always visually distinct, necessitating a merge to balance detectability and informativeness. Below we provide a step-by-step description of the label-augmentation procedure used to subdivide the `Built Area' class of ESRI LULC into Urban and Rural subclasses using GHS-SMOD.

\begin{enumerate}
    \item Both datasets were projected to WGS84 and spatially aligned at 10~m resolution in GEE. The GHS-SMOD layer (1~km) was resampled via nearest-neighbor interpolation to the ESRI LULC grid to preserve categorical integrity.
    \item Label augmentation was applied only to pixels classified as `Built Area' in ESRI LULC. Non-built classes (e.g., Crops, Water, Forest) were retained as-is.
    \item GHS-SMOD defines eight settlement classes based on population density and built-up surface share. These were grouped into two superclasses following the degree of urbanization (DoU) framework recommended by the European Commission (Table~\ref{tab:smod_scheme}). 
\end{enumerate}

\begin{figure}[!ht]
    \centering
    \includegraphics[width = 0.90\textwidth]{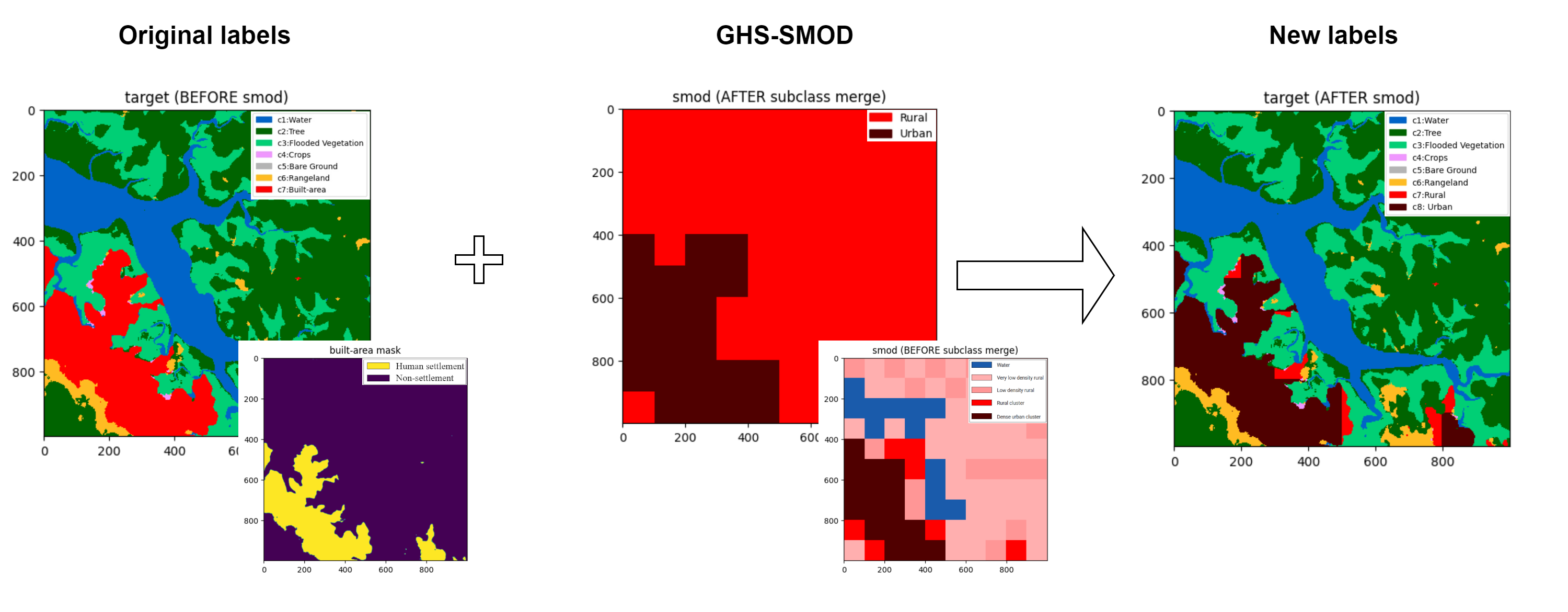}
    \caption{A demonstration of the label augmentation preprocessing, applied to one tile. Information from ESRI LULC and GHS-SMOD is combined to enhance the `Built Area' class. 
    \textbf{Left}: The original labels, with bright red regions denoting `Built Area'---only these regions are affected by the label augmentation.
    \textbf{Center}: The GHS-SMOD labels for this tile, with dark red being urban and bright red being rural. The small tile shows the original eight-class GHS-SMOD dataset, before it was merged into an urban and rural dataset (large tile).
    \textbf{Right}: The resulting labels for this tile after label augmentation preprocessing.}
    \label{fig:label_aug}
\end{figure}

The chosen configuration aggregates GHS-SMOD classes into two broad categories: `urban' and `rural.' This approach consolidates the more detailed subclasses of both rural and urban into their respective superclasses. A visualization of this label augmentation applied to a sample tile is shown in Figure~\ref{fig:label_aug}. The final set of target labels after preprocessing is summarized in Table~\ref{tab:classes_after_preprocessing}.

\begin{table}[!ht]
\centering
\begin{tabular}{|l|l|l|l|}
\hline
& \textbf{Class} & \textbf{Label}\\
\hline
1 & Water & 0\\
2 & Trees & 1\\
3 & Flooded Vegetation & 2\\
4 & Crops & 3\\
5 & Bare Ground & 4\\
6 & Rangeland & 5\\
7 & Rural & 6\\
8 & Urban & 7\\
9 & Missing + Snow/Ice + Clouds & -1\\
\hline
\end{tabular}
\caption{Target labels after preprocessing (-1 denotes an ignored class)}
\label{tab:classes_after_preprocessing}
\end{table}  

\subsubsection*{Input data}
As described earlier, data is saved in tiles of dimension 10~km $\times$ 10~km (1000~px $\times$ 1000~px). This differs from the input dimensions used by the model. In this study, an input size of 250~px $\times$ 250~px was deemed more suitable, as one tile of size 1000~px $\times$ 1000~px can be evenly split into 16 sub-tiles of this size. Figure~\ref{fig:data_gathering_model_input}(c) visualizes the data throughput process.

Preprocessing the data is a crucial step in the training process, as it involves normalizing input data, handling missing data, and remapping labels. This step ensures that data is consistently and correctly formatted for input into the deep model during training. First, the satellite image bands are normalized to values between zero and one. Next, pixels with missing data are addressed. During the data retrieval step, a pixel quality mask was applied to filter out pixels affected by clouds or shadows, marking all pixels without valid observations with the value negative infinity to indicate missing data. Leaving these pixels as is would disrupt the model throughput, while removing each tile containing at least one missing pixel would result in unnecessary data loss. To address this issue, these pixels are assigned the value zero. The final step of preprocessing involves remapping the target labels, which is conducted in three substeps: re-indexing the labels to a value between 0 and \( N-1 \) (where \( N \) is the number of classes), removing unwanted classes, and augmenting class labels by merging different label datasets. Class removal, i.e., marking a class to be ignored, is applied to two classes: Snow/Ice and Clouds. Snow/Ice is a very limited land cover class on the African continent and would therefore risk needlessly interfering with other class predictions. Clouds represent an unknown land cover class due to noisy data and are thus excluded, as this setup utilizes a different satellite data collection.

\subsubsection*{Training a deep model} 

The earth surface contains diverse landscapes, including objects at multiple scales such as forests, water bodies, urban areas, and agricultural regions. Additionally, it includes irregularly shaped objects like rivers, coastlines, and vegetation boundaries. Therefore, multi-scale feature extraction is essential for accurate segmentation. To address this complexity, we adopt the DeepLabV3 semantic segmentation architecture, which uses Atrous Spatial Pyramid Pooling (ASPP) to capture contextual information across multiple receptive fields and effectively delineate irregular settlement boundaries \cite{chen2017rethinking}. DeepLabV3’s dilated convolutions allow the model to capture long-range spatial dependencies more efficiently than traditional encoder-decoder networks such as U-Net or SegNet \cite{ronneberger2015u,badrinarayanan2017segnet}. For the backbone of our deep learning model, we employ ResNet-50 which offers a strong balance between representational depth and computational efficiency \cite{he2016deep}. This configuration is computationally feasible for deployment on high-performance systems (such as the National Academic Infrastructure for Supercomputing in Sweden, NAISS) and for large-area processing in GEE.

Our methodological hypothesis is that combining a deep semantic segmentation model with multi-source satellite data (Landsat-8, VIIRS, ESRI LULC, GHS-SMOD) enables the production of high-resolution, spatially consistent urban-rural maps that outperform existing coarse-resolution datasets. To ensure a fully independent evaluation, model performance is assessed using the DHS dataset, which was not used in training. Accuracy was computed across 20 multiple imputations of DHS urban-rural labels, providing robust statistical estimates of classification performance.

\begin{figure}[!ht]
    \centering
    \includegraphics[width = 0.95\textwidth]{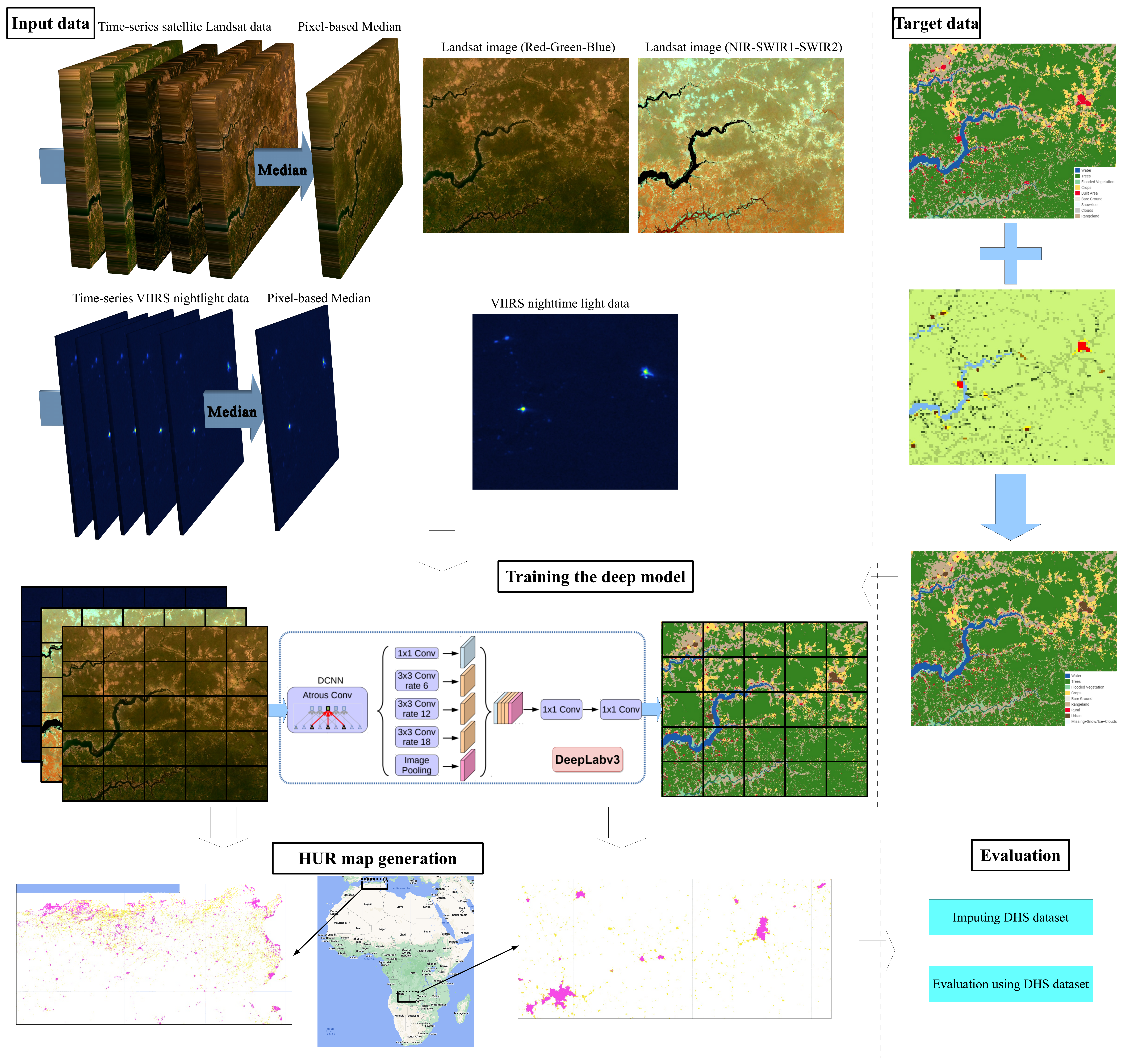}
    \caption{A schematic of the methodology used to generate our high-resolution urban-rural map. In the first step (``Input data'') the satellite imagery---Landsat 8 and VIIRS nighttime light---used to train our model are processed. In the second step (``Target data''), the ESRI LULC is integrated with rural and urban classifications to create the target dataset. The third step (``Training the deep model'') utilizes both the input and target data to train our deep learning model. The trained model generates a high-resolution map at a continental scale (step four, ``HUR map generation''), which is then evaluated both at a country level and annually over time (step five, ``Evaluation'').}
    \label{fig:flowchart}
\end{figure}

The model configuration employs DeepLabV3 as the classification core and bases its predictions on input data from each year. This setup allows the model to produce land cover estimates on a year-by-year basis. Specifically, the model is trained on satellite data from the year 2020, with labels sourced from the ESRI LULC (2020) dataset and GHS-SMOD (2020). The training setup is depicted in Figure~\ref{fig:flowchart}. The architecture follows a standard DeepLabV3 model, with ResNet-50 serving as its backbone. This configuration leverages the strengths of DeepLabV3 in capturing complex features at multiple scales and the robust feature extraction capabilities of ResNet-50.

The loss function used to train the model is cross-entropy loss, which is commonly used for classification tasks and measures the differences between two probability distributions. It is applied to the model output on a pixel-level. For each pixel in the input image, a vector of logits is given as output by the model, which after application of softmax can be viewed as a probability distribution over all classes. The definition of the cross-entropy loss function, as applied to each pixel is shown in Equation~\ref{eq:Loss}, where $y$ is a one-hot encoded vector of the target label, $\hat{y}$ the vector containing the class probabilities, $K=8$ is the number of classes, and $k$ is used as an index for the classes. The loss value is averaged over the pixels in the minibatch.

\begin{equation}
    L(\hat{y},y)=-\sum^K_{k=1} y^{(k)} log(\hat{y}^{(k)})
    \label{eq:Loss}
\end{equation}

The model was trained for a fixed total of 100 epochs using the Adam optimizer with a constant learning rate of $1 \times 10^{-5}$. Training and validation employed a cross-entropy loss function, with class-specific weighting derived from the inverse frequency of each class when country-level class distributions were available. Although no explicit early-stopping criterion was implemented, model selection was performed via continuous monitoring of the validation loss, and the checkpoint corresponding to the lowest validation loss across all epochs was retained as the best model. Training was implemented in PyTorch on NAISS infrastructure. The best-performing model (minimum validation loss) was used for final inference.

\subsubsection*{Handling class imbalance: Weighted loss}
\begin{figure}[!ht]
    \centering
    \includegraphics[width = 0.6\textwidth]{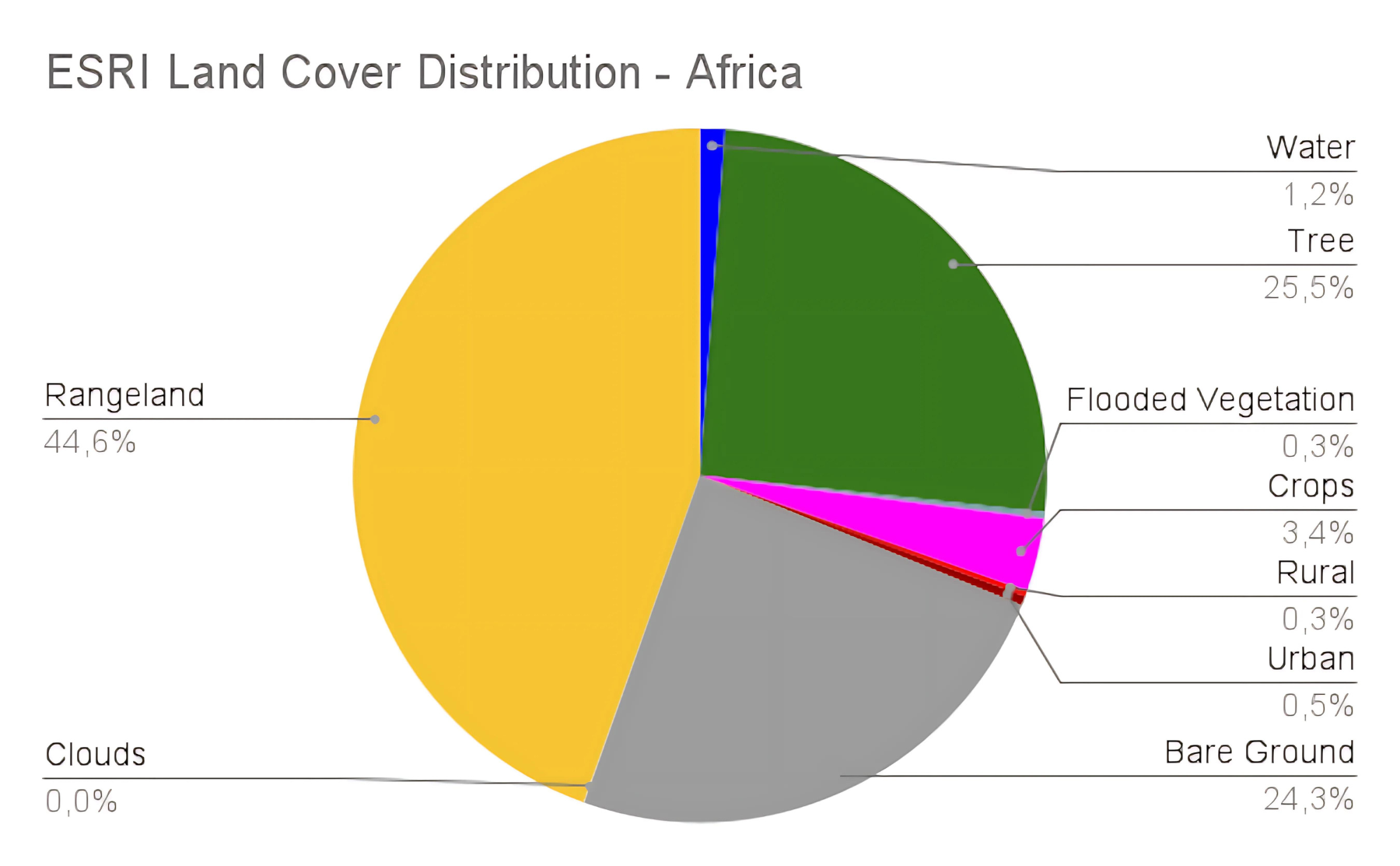}
    \caption{The distribution of land cover classes across Africa in the ESRI LULC 2020 map. }
    \label{fig:class_dist}
\end{figure}

Achieving a balanced class distribution in the training set is crucial to avoid biases toward more prevalent classes. However, achieving this balance is often challenging, especially in cases where some classes are significantly underrepresented compared to others. This is particularly evident in land cover classification, where larger classes can be more than ten times the size of smaller ones. Figure~\ref{fig:class_dist} illustrates the class distribution of ESRI LULC (2020) for the African continent, highlighting the imbalance. To address class imbalance, various strategies can be employed, with weighted loss being one of the most effective. Weighted loss modifies the loss function to assign greater weight to underrepresented classes, thereby increasing the model's focus on these classes and penalizing misclassifications more heavily. This approach helps in improving the model's performance on minority classes.

Several weighting strategies in Equation~\ref{eq:WeightedLoss} were explored: complement probability \((w_{k} = 1-p_{k})\), negative log of the probability \((w_{k} = -log(p_{k}))\), and inverse probability \((w_{k} = 1/p_{k})\). Among these, inverse probability weighting was chosen due to its superior performance in capturing rural settlements. This choice was based on visual inspections and evaluation score comparisons, which indicated that inverse probability weighting provided the best balance between detecting rare classes and overall classification accuracy.

\begin{equation}
    L(\hat{y},y)=-\sum^K_{k=1} w_{k} y^{(k)} log(\hat{y}^{(k)})
    \label{eq:WeightedLoss}
\end{equation}

\subsubsection*{Five-fold cross-validation for intermediate class labels}
An effective evaluation strategy is essential for assessing how well a model generalizes to unseen data. Five-fold cross-validation is a robust method for this purpose, as it provides insight into the model's performance across different subsets of the data. In this approach, the dataset is divided into five folds, and multiple models are trained on various fold configurations. Each fold serves as the test set exactly once, while the remaining folds are used for training and validation. For evaluating the performance of the deep model in this study, a five-fold cross-validation strategy was employed. This involved training five distinct models, each on a different fold configuration. Specifically, each configuration consists of three folds used for training, one fold for validation, and one fold for testing, denoted as \(\{(Train, Validation, Test)\}\). The full set of fold configurations used for model training was: \(\{(123, 4, 5), (234, 5, 1), (345, 1, 2), (451, 2, 3), (512, 3, 4)\}\) In this setup, each model was trained on a unique combination of folds, with the corresponding validation and test folds ensuring a comprehensive evaluation across different subsets of the data.

Spatial autocorrelation is a fundamental aspect of satellite imagery \cite{spatial_dependency_2022}, where neighboring pixels often exhibit higher similarity compared to those that are farther apart. This spatial dependence can influence the performance of machine learning models trained on such data. When a model is trained on one image tile and evaluated on an adjacent tile, the proximity of the tiles may result in an overestimation of the model’s generalization capabilities, as the data may not be truly independent. This challenge underscores the importance of careful consideration in designing the train-test split to avoid skewed performance metrics. Standard random fold splitting, which typically ignores spatial dependencies, can lead to artificially inflated performance results because adjacent or nearby tiles may share similar features. Rolf et al. \cite{spatial_dep_checkerboard} explored the impact of spatial separation on model performance by using a checkerboard grid with varying spacing. Their findings suggested that tighter grid layouts, which maintain spatial separation, often yielded more accurate assessments of model performance compared to more loosely spaced grids, indicating a potential overestimation in performance when spatial dependencies are not considered. To address this issue, the study adopts a country-wise partitioning approach for the data split. By assigning entire countries to each fold, this method effectively enforces spatial boundaries and minimizes the risk of data leakage between training and test sets. Additionally, country-wise partitioning allows for evaluation at the country level, providing a more realistic measure of the model's performance across diverse geographical regions.

Data was divided into five folds, with each fold comprising a subset of countries. To achieve balanced folds, countries were first sorted by area and then assigned to the folds in a cyclic manner. This approach ensured that each fold contained a roughly equal number of countries and was of similar size. The resulting folds and their corresponding countries are as follows:

\textit{Fold 1}: Algeria, Niger, Mauritania, Mozambique, Central African Republic, Zimbabwe, Guinea, Malawi, Togo  

\textit{Fold 2}: Democratic Republic of the Congo, Angola, Egypt, Zambia, Madagascar, Congo, Ghana, Eritrea, Guinea-Bissau, 

\textit{Fold 3}: Sudan, Mali, United Republic of Tanzania, Morocco, Botswana, Côte d'Ivoire, Uganda, Benin, Lesotho 

\textit{Fold 4}: Libya, South Africa, Nigeria, South Sudan, Kenya, Burkina Faso, Senegal, Liberia, Equatorial Guinea 

\textit{Fold 5}: Chad, Ethiopia, Namibia, Somalia, Cameroon, Gabon, Tunisia, Sierra Leone, Burundi

\begin{figure}[!ht]
    \centering
    \includegraphics[width = 0.45\textwidth]{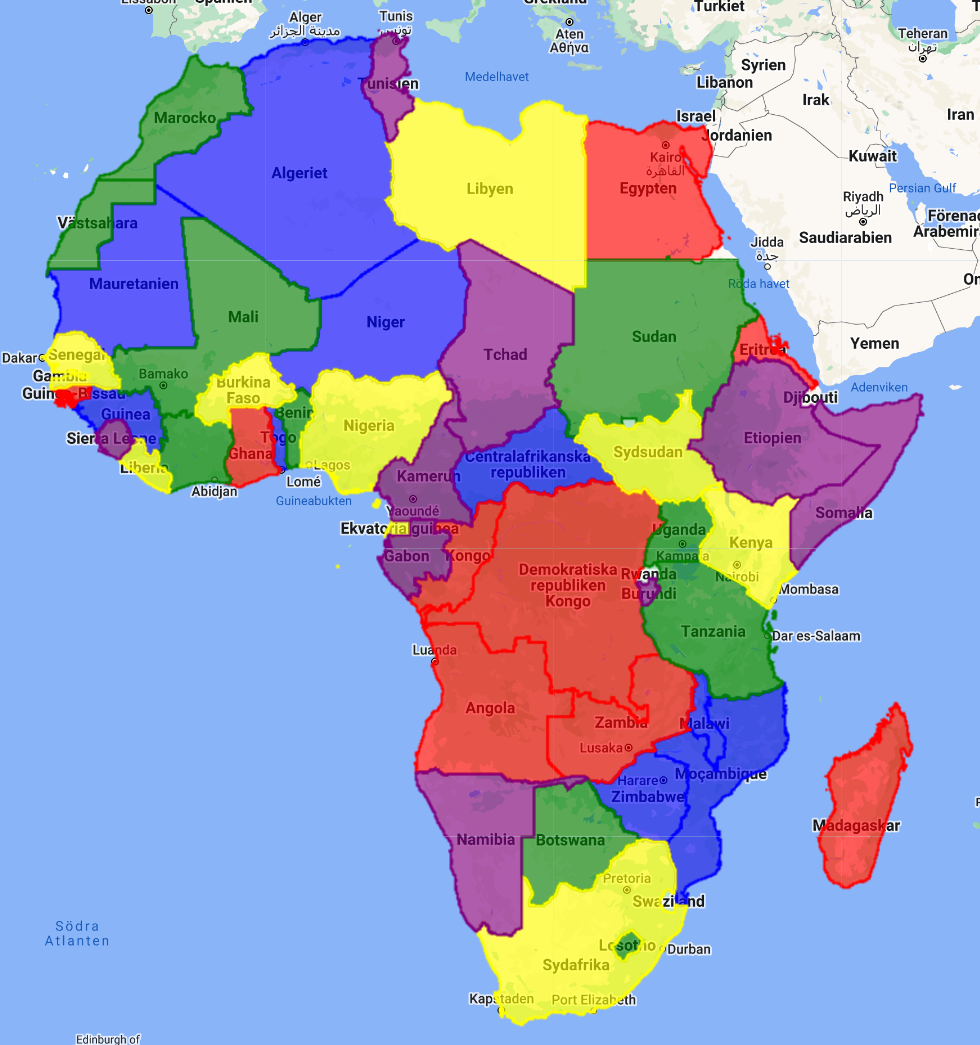}
    \caption{The five folds in which countries are split, highlighted in different colors (fold one: blue, fold two: red, fold three: green, fold four: yellow, and fold five: purple). Screenshot taken from the GEE.}
    \label{fig:cross_fold}
\end{figure}

A visualization of the split is displayed in Figure~\ref{fig:cross_fold}, with each fold highlighted by a separate color.

\subsubsection*{Output generation: Smooth tiling}
After training and evaluation, the models are ready for deployment. They are applied to generate an urban-rural map of Africa based on the predicted land cover classes. We have released the urban-rural map, which has been independently evaluated using DHS survey data. It is important to note that the five-fold cross-validation described earlier was used only for training and evaluating the intermediate land cover classifications. Evaluating the model on its own prediction maps does not constitute a true assessment. Therefore, the final outputs are generated and categorized into three classes: urban, rural, and NonHS.

\begin{figure}[!ht]
    \centering
    \begin{subfigure}{.3\textwidth}
      \centering
      \includegraphics[width=.98\linewidth ]{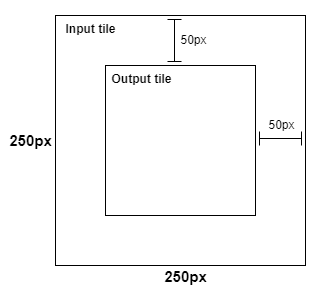}
      \caption{ }
    \end{subfigure}%
    \begin{subfigure}{.6\textwidth}
      \centering
      \includegraphics[width=.98\linewidth ]{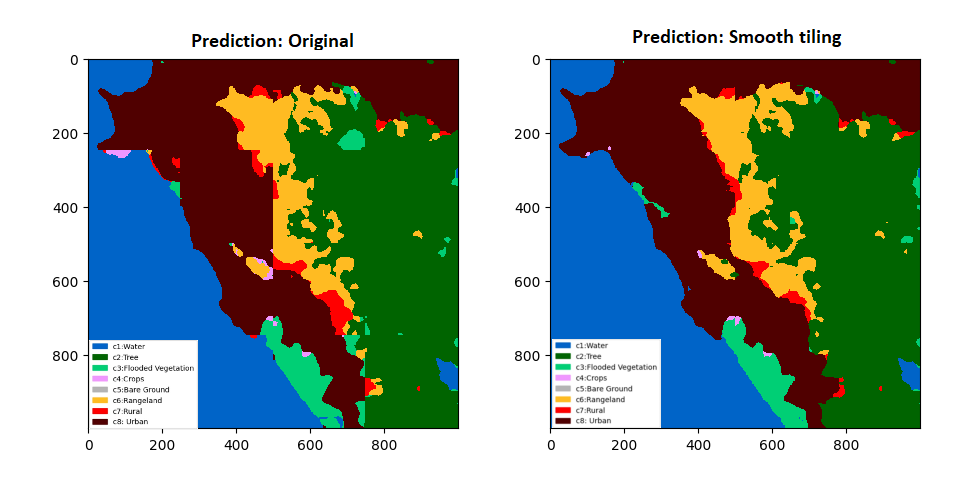}
      \caption{ }
    \end{subfigure}
    \caption{An illustration of the smooth tiling output generation process. \textbf{(a)} The input and output dimensions of our tiling scheme. Inference is done over the entire image (large square) but only the center (smaller square) is used to generate our urban-rural map. \textbf{(b)} A comparison between the original predictions (left) and the predictions with tile smoothing (right). In the original image, clear edge artifacts are visible, while the smoothed image is more cohesive.}
    \label{fig:smooth_tiling_unsmooth_smoothened}
\end{figure}

The simplest procedure would be to feed to the model, as input, a grid of non-overlapping tiles covering the desired area and join them together in one larger image. This method does however lead to unwanted edge effects in the final image. As the models have been trained on images of dimensions 250~px $\times$ 250~px, using the same input dimensions when running inference is the natural choice. The deep model can be applied to images of arbitrary size, as it is a fully-convolutional-network, but should work optimally for the dimensions trained on. A method for smoother image tiling was implemented \cite{pfeuffer2019semantic} to reduce the aforementioned edge effects. The edge effects are a result of the outer edge of an image getting less surrounding context, leading to neighboring tile edges not always matching. To reduce this problem, only the center of the prediction (which has the most context) is kept, while the outer edge of the prediction is thrown away, as shown in Figure~\ref{fig:smooth_tiling_unsmooth_smoothened}(a). A larger number of (partially overlapping) image tiles are then needed to stitch together the full output tile. With neighboring tiles now receiving more context along the edges, the resulting urban-rural map appears smoother and more accurate, as shown in Figure~\ref{fig:smooth_tiling_unsmooth_smoothened}(b).

To summarize, we employed a cropping-based overlap strategy to geerate the final map using the steps below,
\begin{itemize}
    \item The full mosaic was divided into overlapping tiles with 50\% overlap between adjacent tiles in both x and y directions. This ensured that each location was covered by multiple inference windows.
    \item From each predicted tile, only the central region with 125~px $\times$ 125~px was retained. The outer border areas, where contextual information is incomplete, were discarded.
    \item The cropped central patches were then mosaicked seamlessly by placing them in their corresponding spatial locations. Therefore, no pixel averaging or blending was performed, and overlapping predictions were replaced by the central crop of each tile.    
\end{itemize}

\section*{Results}

Results are provided in this section and performance is evaluated using rural and urban labels from the DHS dataset as detailed earlier as reference labels. 

\subsection*{DHS-based evaluation}

According to DHS guidelines, the DHS program does not itself classify survey locations as urban or rural. Instead, it relies on classifications provided by national governments or statistical offices. These definitions are country-specific and may vary substantially across countries. In our work, we use DHS survey clusters as validation points for settlement classification, but without re-defining their urban-rural status. This approach avoids the challenge of reconciling multiple national definitions, which is beyond the scope of this study.

For the urban and rural validation data, we use perturbed DHS cluster points collected after 2013 and impute their plausible unperturbed locations. For each cluster, we generate imputed locations by drawing from a Bayesian posterior distribution. The posterior is constructed from a uniform prior on human-settlement pixels in the JRC Human Settlement (HS) map, combined with a likelihood derived from the DHS perturbation procedure.

For the non-human settlement (NonHS) validation data, we approximate locations using DHS-perturbed points that fall outside settlement areas in the JRC HS map. Because DHS clusters are representative of towns, cities, and villages, and the perturbation is applied uniformly at random, these points are approximately uniformly distributed near settlement boundaries, making them suitable for evaluating NonHS classifications.

\subsection*{Evaluation metrics}
To quantitatively assess the performance of the generated urban-rural maps, we compute accuracy, precision, recall, and the Cohen's kappa coefficient using the DHS dataset as an independent reference. All metrics are derived from the confusion matrix elements---true positive (TP), false positive (FP), true negative (TN) and false negative (FN)---which are computed separately for the Non-HS, Rural, and Urban classes. In this three-class classification, these values are computed independently for each class \(c\in \{ \text{Non-HS}, \text{Rural}, \text{Urban} \} \) using the one-vs-rest strategy:
\begin{itemize}
    \item TP is the number of samples correctly predicted as belonging to class $c$.
    \item FP is the number of samples incorrectly predicted for class $c$, while their true label belongs to another class.
    \item TN is the number of samples that are neither predicted as class $c$ nor actually belong to class $c$.
    \item FN is the number of samples belonging to class $c$ that were misclassified as another class.
\end{itemize}

The evaluation metrics include accuracy, precision, recall, and Cohen's kappa coefficient \cite{cohen1960coefficient}. Accuracy measures the proportion of correctly classified samples across all classes displayed in Equation~\ref{eq:Accuracy}. Precision quantifies how many predictions of a given class are correct. For each class $c$ precision is calculated according to Equation~\ref{eq:Precision}. Recall measures how many actual samples of a given class were correctly identified according to Equation~\ref{eq:Recall}. Finally, Cohen's kappa coefficient evaluates classification agreement between the predicted map and the DHS reference labels while correcting for chance agreement. It is defined in Equation~\ref{eq:Kappa}.

\begin{equation}
    \text{Accuracy} = \frac{TP + TN}{TP + FP + FN + TN}
    \label{eq:Accuracy}
\end{equation}

\begin{equation}
    \text{Precision}_{c} = \frac{TP{c}}{TP_{c} + FP_{c}}
    \label{eq:Precision}
\end{equation}

\begin{equation}
    \text{Recall}_{c} = \frac{TP{c}}{TP_{c} + FN_{c}}
    \label{eq:Recall}
\end{equation}

\begin{equation}
\begin{aligned}
\text{Cohen's kappa coefficient } \kappa &= \frac{p_o - p_e}{1 - p_e}, \text{ where} \\
p_o &= \frac{1}{N} \sum_{c} TP_{c}, \text{ and}\\
p_e &= \sum_{c} \left( \frac{TP_{c} + FP_{c}}{N} \right)
               \left( \frac{TP_{c} + FN_{c}}{N} \right).
\end{aligned}
\label{eq:Kappa}
\end{equation}

\subsection*{Quantitative evaluation with the DHS}

\begin{figure}[!ht]
    \centering
    \begin{subfigure}{.98\textwidth}
      \centering
      \includegraphics[width=.98\linewidth ]{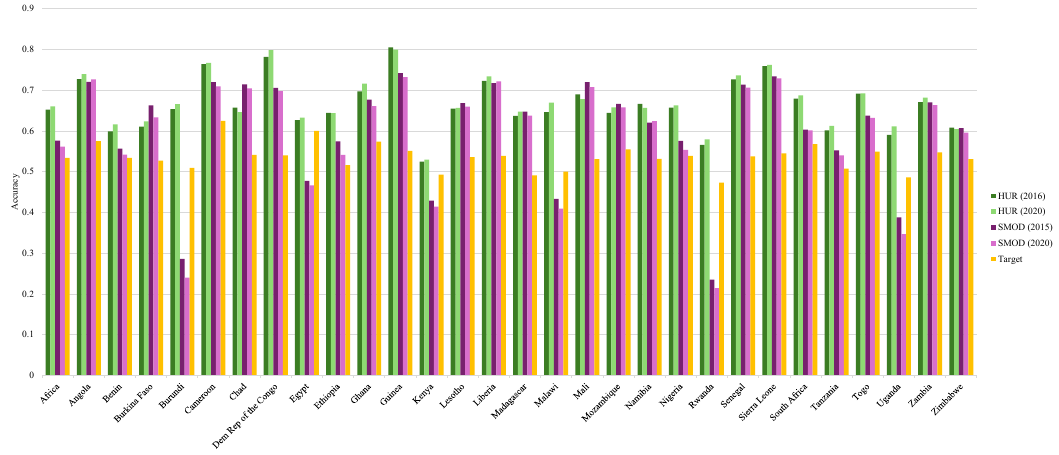}
      \caption{ }
    \end{subfigure}
    \begin{subfigure}{.98\textwidth}
      \centering
      \includegraphics[width=.98\linewidth ]{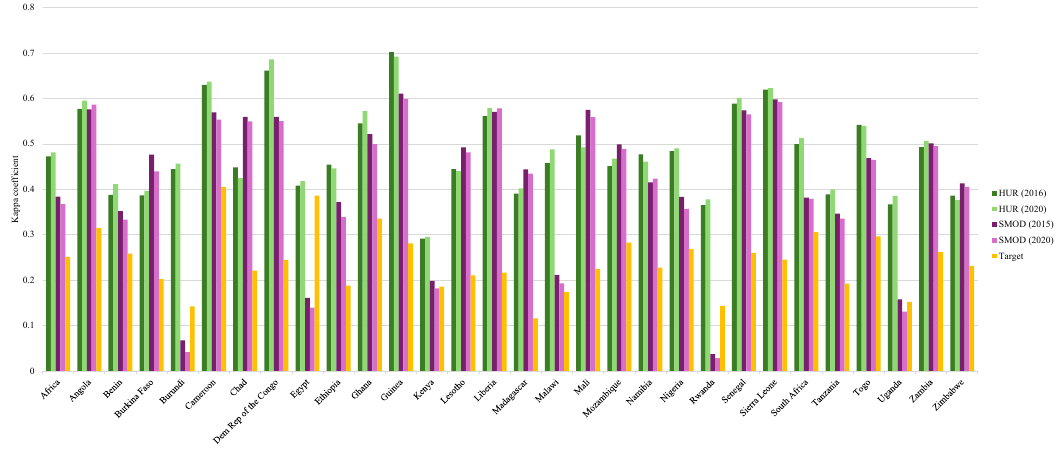}
      \caption{ }
    \end{subfigure}
    \caption{A country-wise comparison of two HUR maps (2016 and 2020), two GHS SMOD maps (2015 and 2020), and the target dataset, evaluated against DHS classification. Two different evaluation metrics are displayed: \textbf{(a)} accuracy, and \textbf{(b)} Cohen's kappa coefficient.
    }
    \label{fig:DHSeval}
\end{figure}

To assess the reliability of our HUR maps, we compared them against the JRC Settlement Model (SMOD) and the synthetic target dataset, using DHS-based validation points across multiple African countries. Two standard metrics including Accuracy and the Kappa coefficient are used in Figure~\ref{fig:DHSeval}.

The evaluation shows consistent improvements for the HUR maps over SMOD. For example, for Guinea in 2020, HUR achieved an accuracy of $0.81$ and a Kappa coefficient of $0.70$, compared to $0.74$ accuracy and $0.60$ Kappa for SMOD. In Burundi, both HUR (2016) and HUR (2020) exceeded 0.65 accuracy, while SMOD remained below 0.30. In Nigeria, HUR (2020) surpassed 0.65 accuracy, whereas SMOD fell below 0.58.

The target dataset, which combines ESRI LULC and SMOD-derived settlement labels, consistently fell below SMOD in both accuracy and Kappa. This reflects the simple combination of datasets can be considered as intermediate training resource rather than a new map.

HUR maps consistently outperform SMOD, particularly in countries where SMOD accuracy is low—such as Burundi, Egypt, Malawi, Rwanda, and Uganda. This demonstrates the robustness of deep learning models in generating reliable rural-urban classifications at continental scale, with consistent improvements across diverse national contexts.

\subsection*{Continental-scale results}

At the continental scale, HUR maps again outperformed SMOD. HUR2016 and HUR2020 reached kappa values of 0.47–0.48 and accuracies of 0.65–0.66, compared to 0.37–0.38 kappa and 0.56–0.57 accuracy for SMOD. The target dataset performed worst (kappa 0.25, accuracy 0.53). Class-specific evaluation are summarized in Table~\ref{tab:precision_recall}.

\begin{table}[!ht]
\begin{center}
\begin{tabular}{c|c|c|c c c | c c c }
\multirow{2}{*}{\textbf{Dataset}} & \multirow{2}{*}{\textbf{Kappa}} & \multirow{2}{*}{\textbf{Accuracy}} & \multicolumn{3}{c|}{\textbf{Precision}} &  \multicolumn{3}{c}{\textbf{Recall}}   \\
  &  &  & Non-HS & Rural & Urban  &  Non-HS & Rural & Urban \\
\hline
Target  & 0.25 & 0.53 & 0.495 & 0.364 & \textbf{0.67}  &  \textbf{0.863} & 0.02 & 0.697 \\
HUR2016  & 0.47 & 0.65 & 0.693 & \textbf{0.757} & 0.543  &  0.766 & 0.374 & 0.851 \\
HUR2020  & \textbf{0.48} & \textbf{0.66} & 0.685 & \textbf{0.757} & 0.568  &  0.79 & 0.379 & 0.833 \\
SMOD2015  & 0.38 & 0.57 & 0.79 & 0.58 & 0.46  &  0.45 & \textbf{0.509} & 0.911 \\
SMOD2020  & 0.37 & 0.56 & \textbf{0.812} & 0.574 & 0.439  &  0.421 & 0.496 & \textbf{0.921} \\
\hline
\end{tabular}%
\caption{An evaluation of five different urban-rural maps against DHS classifications on four metrics: precision, recall, Cohen's kappa coefficient, and accuracy.}
\label{tab:precision_recall}
\end{center}
\end{table}

For Non-HS class, SMOD2020 achieved the highest precision (0.812) but had low recall (0.421), indicating under-detection of NonHS areas. HUR2020 showed more balanced performance (precision 0.685, recall 0.79). For Rural class, HUR maps yielded the strongest precision (0.757), while SMOD had higher recall (0.509), suggesting SMOD identifies more rural areas but at the cost of false positives. It can also be due to its coarser spatial resolution. For urban class, SMOD achieved very high recall (0.911–0.921) but low precision (0.439–0.460), reflecting overestimation of urban areas. HUR maps balanced recall (0.833–0.851) with substantially higher precision (0.543–0.568).

These results confirm that while SMOD captures urban areas broadly, its coarse 1~km resolution inflates urban extents and reduces precision. By contrast, HUR’s 10~m resolution better captures rural-urban heterogeneity, leading to improved overall accuracy.

\subsection*{Location uncertainty}

Because DHS cluster coordinates are intentionally displaced to preserve household confidentiality, direct point-based validation introduces spatial uncertainty. To address this, we adopt a multiple-imputation framework using 20 independent realizations of each cluster's possible true location. Let the released DHS dataset $D^{DHS}$ contain cluster points labeled as urban ($D^{Urban}$)  or rural ($D^{Rural}$). DHS applies a displacement of up to 2 km for urban clusters and up to 5 km for rural clusters, meaning the published coordinates represent perturbed rather than true locations.

Let the generated HUR map be denoted as $(M(\cdot))$, which returns a class \(c\in \{ NonHS, Rural, Urban \} \)  for any spatial coordinate. We treat each cluster's unobserved true location as a random point distributed within a disk centered on the released DHS coordinate, with radius determined by the DHS displacement rule. Following DHS documentation, the perturbation is drawn uniformly by distance (not uniformly by area), meaning points closer to the published coordinate have higher probability density.

From this distribution we draw 20 imputed realizations of each cluster location. Each realization yields an independent evaluation of $(M(\cdot))$. Our reported expected accuracy and its uncertainty are therefore based on multiple-imputation evaluation, where final estimates and confidence intervals are pooled across the 20 imputations.

\begin{figure}[!ht]
    \centering
    \begin{subfigure}{.48\textwidth}
      \centering
      \includegraphics[width=.98\linewidth ]{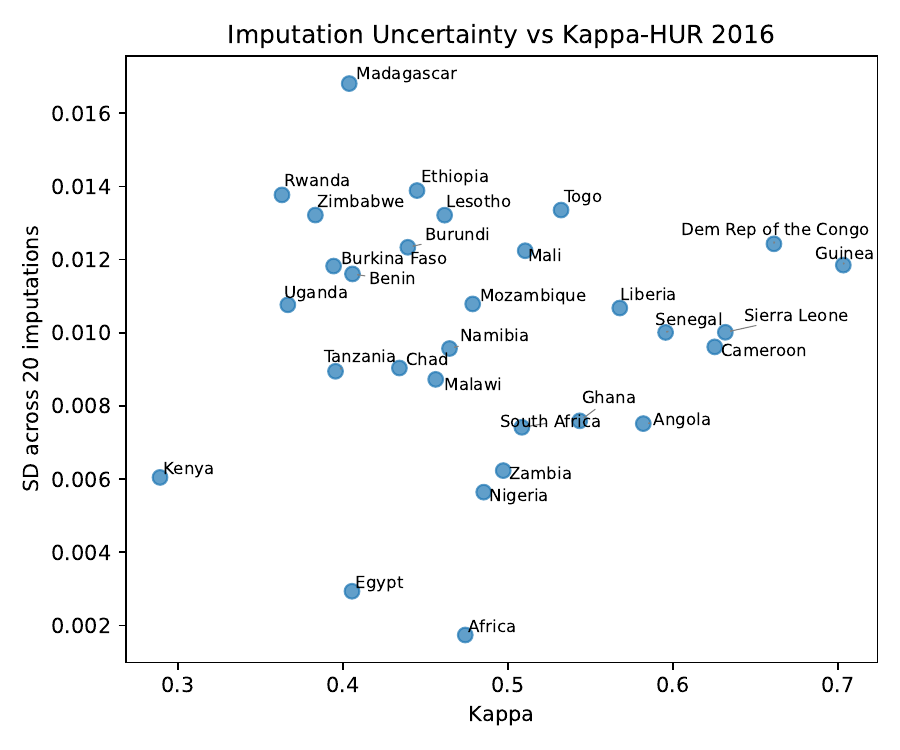}
      \caption{ }
    \end{subfigure}
    \begin{subfigure}{.48\textwidth}
      \centering
      \includegraphics[width=.98\linewidth ]{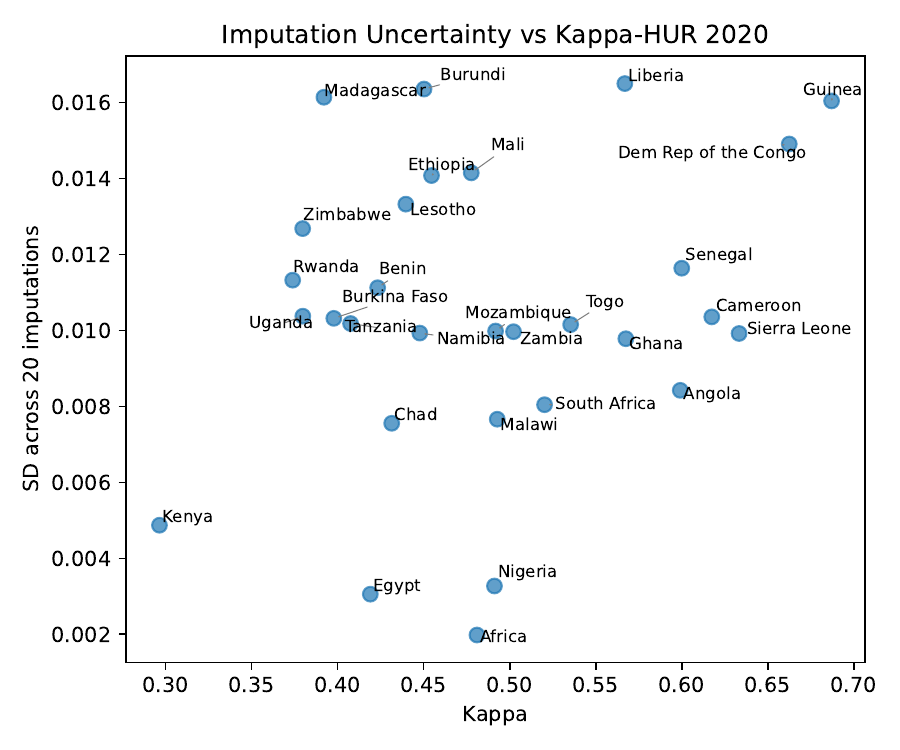}
      \caption{ }
    \end{subfigure}
    \begin{subfigure}{.48\textwidth}
      \centering
      \includegraphics[width=.98\linewidth ]{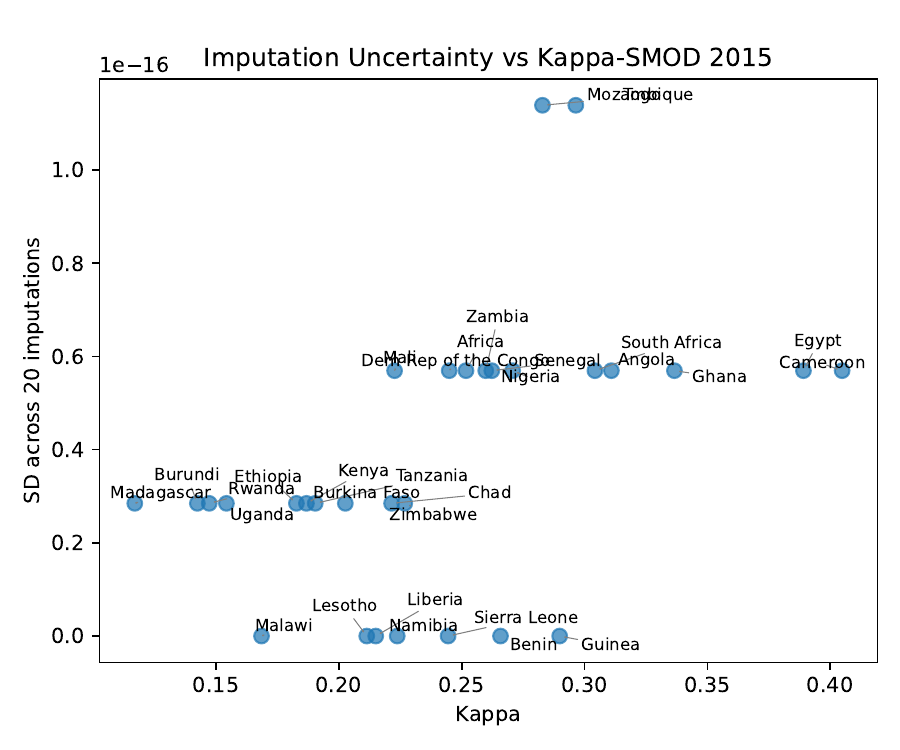}
      \caption{ }
    \end{subfigure}
    \begin{subfigure}{.48\textwidth}
      \centering
      \includegraphics[width=.98\linewidth ]{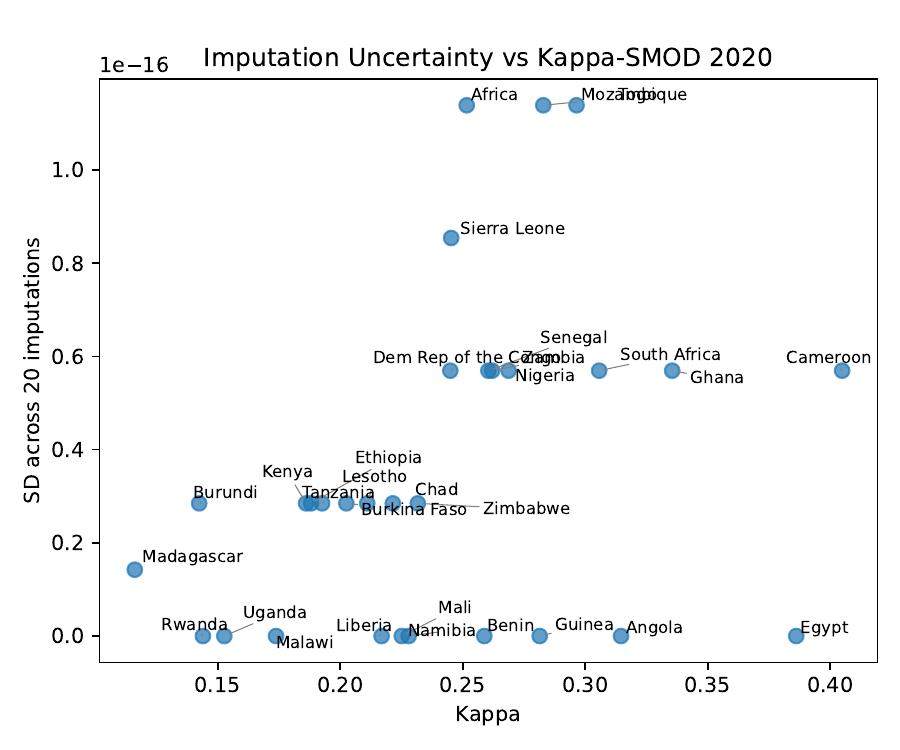}
      \caption{ }
    \end{subfigure}
    \caption{Plots of the average (mean) and the standard deviation of the Cohen's kappa coefficient across 20 imputations of the DHS confidential locations for \textbf{(a)} HUR 2016; \textbf{(b)} HUR 2020; \textbf{(c)} GHS SMOD 2015; and \textbf{(d)} GHS SMOD 2020. (The horizontal bands in the GHS SMOD plots are an artifact of the low resolution of these maps.)}
    \label{fig:KappaSdCountries}
\end{figure}

\begin{figure}[!ht]
    \centering
    \begin{subfigure}{.48\textwidth}
      \centering
      \includegraphics[width=.98\linewidth ]{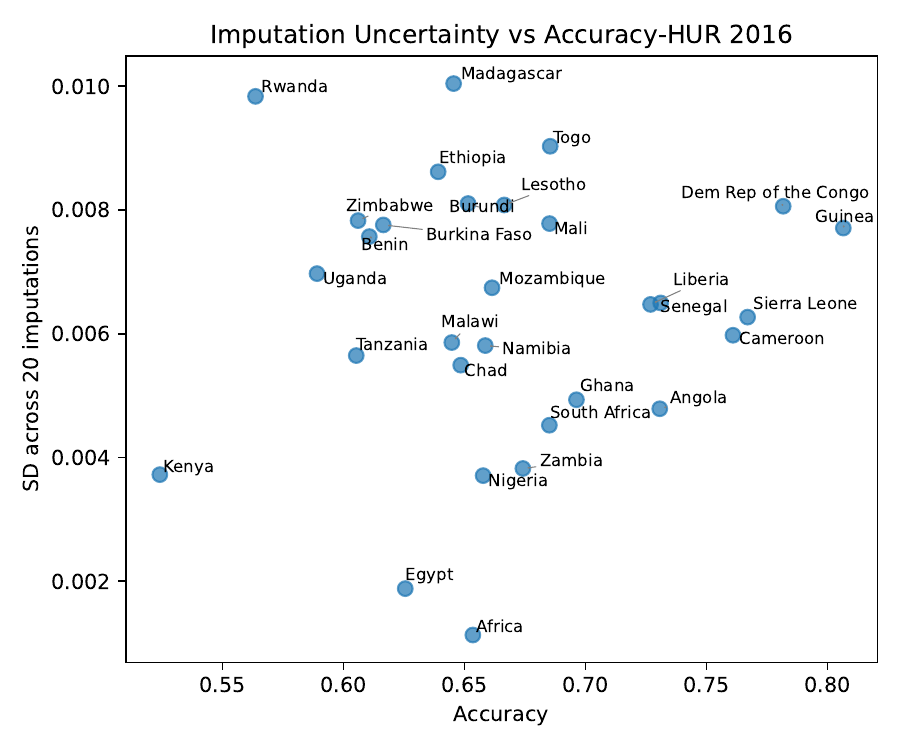}
      \caption{ }
    \end{subfigure}
    \begin{subfigure}{.48\textwidth}
      \centering
      \includegraphics[width=.98\linewidth ]{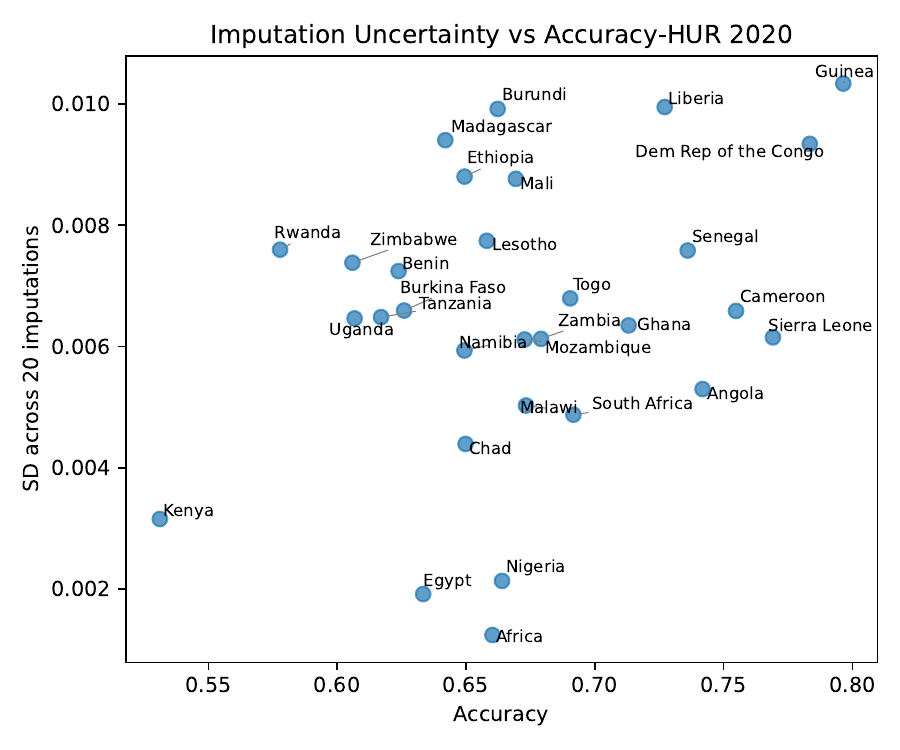}
      \caption{ }
    \end{subfigure}
    \begin{subfigure}{.48\textwidth}
      \centering
      \includegraphics[width=.98\linewidth ]{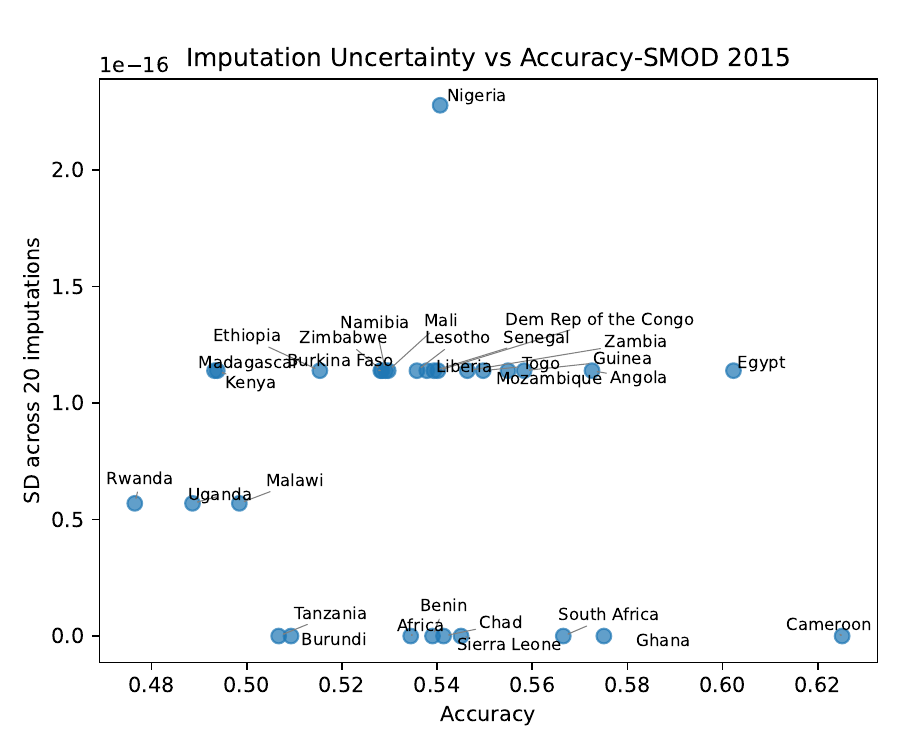}
      \caption{ }
    \end{subfigure}
    \begin{subfigure}{.48\textwidth}
      \centering
      \includegraphics[width=.98\linewidth ]{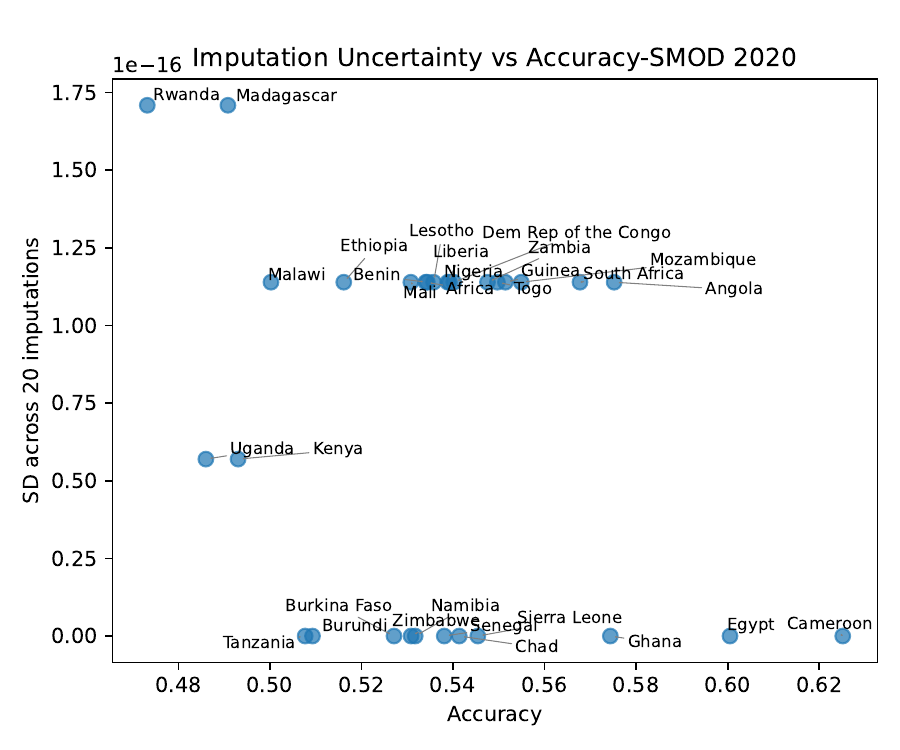}
      \caption{ }
    \end{subfigure}
    \caption{Plots of the average (mean) and the standard deviation of the map's accuracy across 20 imputations of the DHS confidential locations for \textbf{(a)} HUR 2016; \textbf{(b)} HUR 2020; \textbf{(c)} GHS SMOD 2015; and \textbf{(d)} GHS SMOD 2020. (The horizontal bands in the GHS SMOD plots are an artifact of the low resolution of these maps.)}
    \label{fig:AccuracySdCountries}
\end{figure}

Country-level comparisons of Kappa coefficients and accuracies, including their between-imputation variability, for our HUR maps and the GHS-SMOD maps are presented in Figures~\ref{fig:KappaSdCountries} and~\ref{fig:AccuracySdCountries}. Compared to the continental results in Figure~\ref{fig:DHSeval}, these plots add the uncertainty dimension, making it possible to assess both central performance and variability across countries. The SMOD maps exhibit very low between-imputation variation, largely because their coarse spatial resolution produces stable predictions that do not change substantially when imputations shift cluster locations by a few kilometers.

\begin{figure}[!ht]
    \centering
    \includegraphics[width = 1\textwidth]{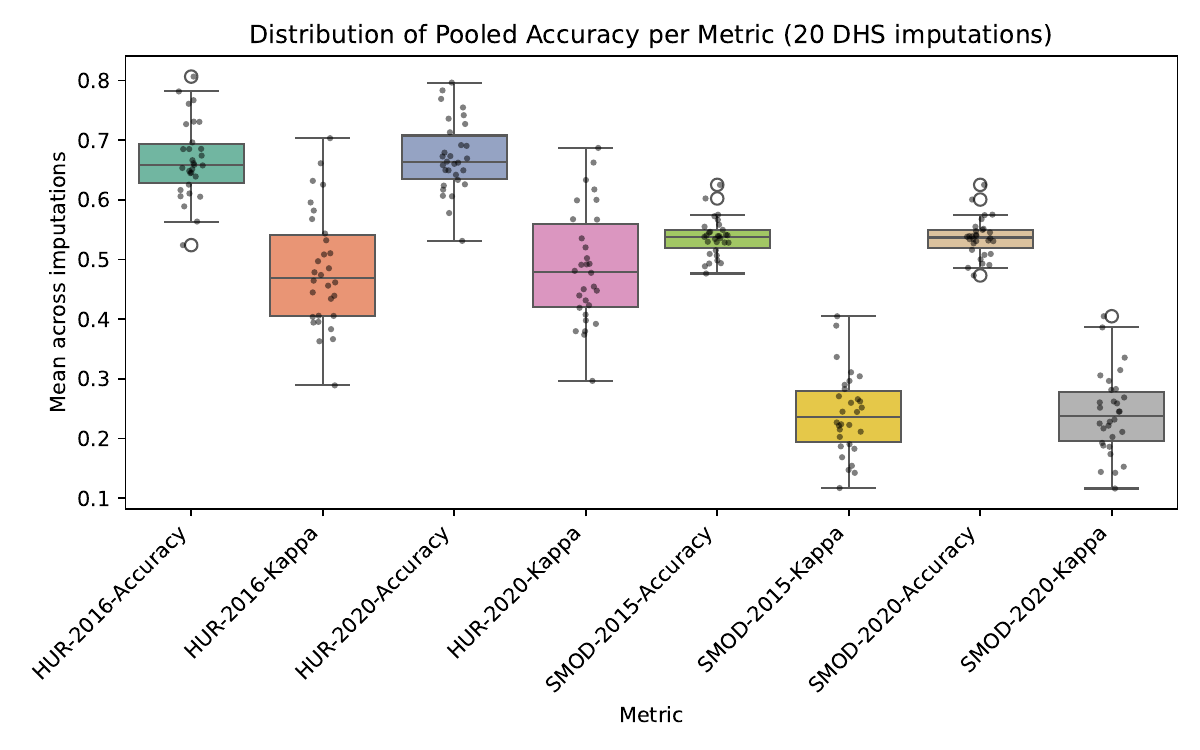}
    \caption{Box and whisker plots of the distribution of the map's accuracy and the Cohen's kappa coefficient across the 20 DHS imputations.}
    \label{fig:MeanSdCountries}
\end{figure}

Figure~\ref{fig:MeanSdCountries} summarizes the pooled mean accuracy and Cohen's kappa coefficient for HUR 2016, HUR 2020, SMOD 2015, and SMOD 2020 across 20 imputations. Although country labels are omitted for clarity, the figure illustrates that the HUR maps achieve both higher mean performance and greater sensitivity to imputation-driven spatial variation. This is an expected outcome: the HUR maps capture fine-scale settlement structure at 10~m resolution, so even small changes in cluster position can affect the classification extracted at that location. In contrast, the much coarser SMOD layers provide spatially smoother representations with limited detail, resulting in nearly identical evaluation outcomes across imputations.

In conclusion, the higher between-imputation variability observed for the HUR maps is a feature of their higher spatial fidelity, not a limitation of the model. High-resolution predictions are naturally more responsive to the uncertainty in displaced survey coordinates, whereas coarse-resolution products such as SMOD remain essentially unchanged under the same imputation framework.

\subsection*{Visual evaluation}

\begin{figure}[!ht]
    \centering
    \includegraphics[width = 1\textwidth]{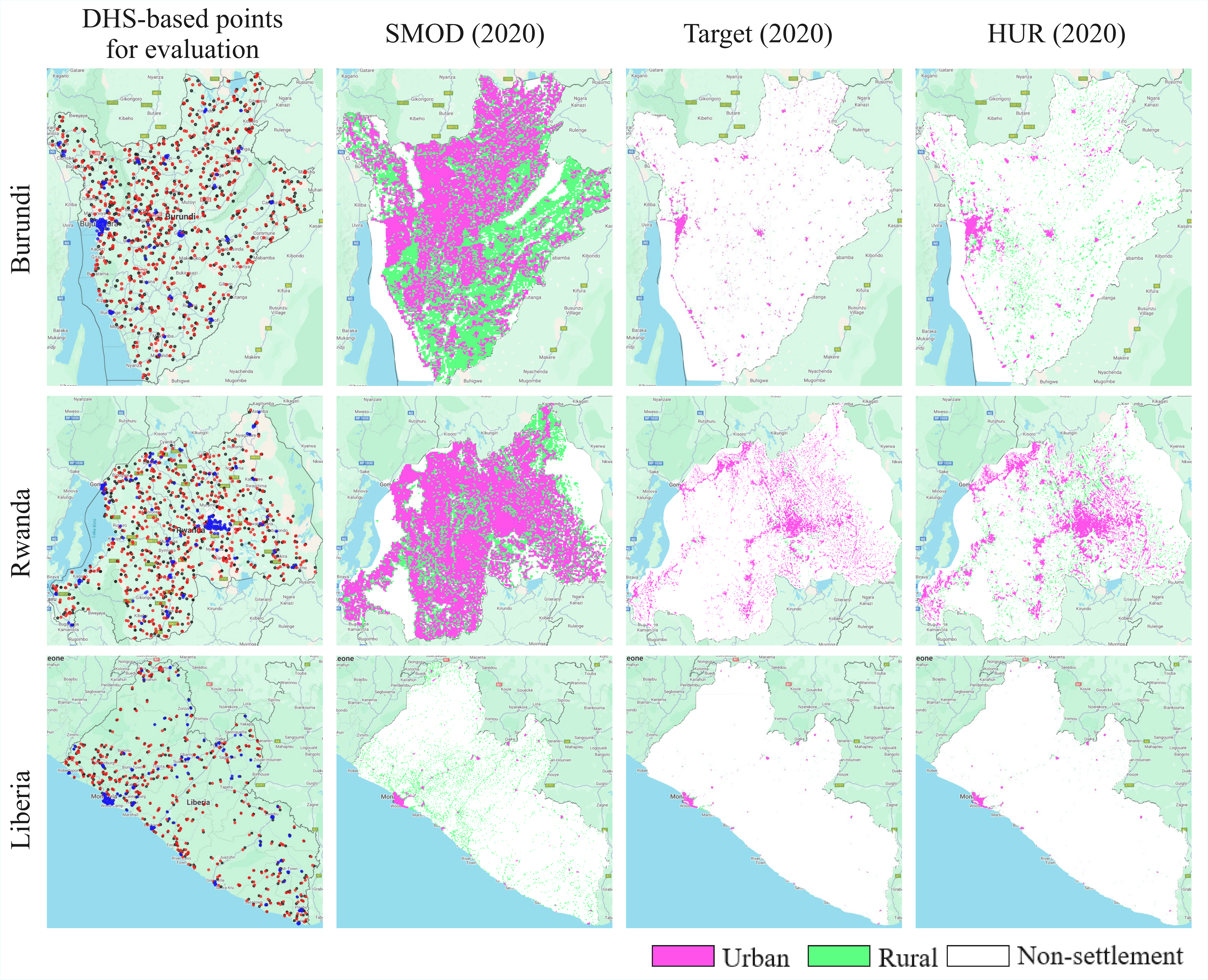}
    \caption{Maps of three countries (Burundi, Rwanda and Liberia) and their DHS survey points (first column), along with the corresponding SMOD 2020 (second column), target (third column), and HUR 2020 (third column) maps. The DHS points are colored black for Non-HS, red for rural and blue for urban classes. In the SMOD, target and HUR maps, NonHS areas are white, rural areas are green, and urban areas are pink.}
    \label{fig:MapsCountries}
\end{figure}

Figure~\ref{fig:MapsCountries} illustrates qualitative comparisons for Burundi, Rwanda, and Liberia.

In Burundi and Rwanda, DHS points reveal densely settled rural landscapes. SMOD (2020), at 1~km resolution, tends to merge these into continuous rural clusters, losing small distinctions. HUR (2020) maintains finer settlement boundaries, distinguishing villages from urban centers. The Target dataset offers intermediate performance, more detailed than SMOD but less precise than HUR.

In Liberia, where settlements are more dispersed, SMOD overestimates rural extents, classifying wide areas as rural. HUR (2020) confines urban areas more closely to the built-up footprint, aligning better with DHS points. Again, the Target dataset falls between the two.

Overall, the visual evidence confirms the quantitative findings. SMOD favors recall but overestimates rural and urban areas, while HUR captures the most realistic rural-urban patterns, with strong alignment to survey-based validation data.

\subsection*{Conclusions}
Overall, the results demonstrate that the HUR maps provide a substantial improvement over existing global products for settlement classification across Africa. Using DHS survey clusters as an independent reference with a principled multiple-imputation framework to account for spatial displacement uncertainty shows that the HUR maps consistently achieve higher accuracy and Kappa coefficients than the widely used GHS-SMOD layers. These performance gains are observed both at the continental scale and across most individual countries, which reflects the model's ability to capture fine-scale rural-urban structure that is not represented at coarser spatial resolutions.

Class-specific evaluation further highlights the advantages of the deep-learning approach. Compared with SMOD, the HUR maps achieve more balanced precision and recall for all three classes (NonHS, Rural, Urban), which particularly reduces the systematic over-estimation of urban areas characteristic of coarse-resolution datasets. The 10~m mapping resolution allows the model to identify small settlements, and heterogeneous rural landscapes that are typically merged or misclassified in lower-resolution products.

The multiple-imputation analysis confirms that the observed improvements remain robust even when DHS location uncertainty is explicitly modeled. As expected, the HUR maps show higher between-imputation variation than SMOD, not due to instability but because fine-resolution predictions are inherently more sensitive to shifts in point location near settlement boundaries. In contrast, the spatially coarse SMOD layers produce nearly identical results across imputations.

Taken together, these findings provide strong empirical evidence that the proposed deep learning framework and leveraging Landsat imagery, ESRI LULC labels, and multi-source contextual information, significantly enhances the accuracy and spatial detail of rural-urban mapping in Africa. The HUR dataset thus offers a more reliable basis for analyzing settlement patterns, monitoring spatial development, and supporting policy-relevant applications at both national and continental scales.

\section*{Discussion}

This study demonstrates the feasibility and advantages of generating high-resolution (10~m) rural-urban maps across Africa using deep learning and multi-source satellite data. Compared with existing global settlement products, particularly the JRC SMOD dataset, the proposed HUR maps achieve consistently higher agreement with independent DHS-based ground truth at both continental and national scales. These improvements are especially pronounced in countries where rural-urban transitions are spatially complex, highly fragmented, or dominated by small and dispersed settlements that are not adequately captured in coarser-resolution datasets.

A central contribution of this work is the demonstration that finer-resolution mapping directly resolves systematic biases observed in kilometer-scale products. SMOD, for example, demonstrated very high recall for urban areas but consistently overestimated urban extent, resulting in low precision. This pattern is expected given that SMOD defines urban areas based largely on population density and built-up intensity, which tend to aggregate diverse settlement types into broad classes. In contrast, the HUR maps showed a more balanced trade-off between recall and precision while preserving the ability to detect both large cities and small rural villages. This result suggests that HUR's deep learning model captures fine-grained spatial signatures of human settlement morphology---such as roof structures, geometric patterns, and contextual land-cover cues---that are unavailable or smoothed out at kilometer-scale resolutions.

Importantly, the higher between-imputation standard deviations observed for the HUR maps should not be interpreted as reduced reliability. Instead, they reflect three expected properties of fine-resolution models: (1) sensitivity to small spatial shifts in DHS cluster locations, particularly near settlement edges; (2) greater responsiveness of detailed spatial surfaces to imputation-based perturbations; and (3) the well-documented phenomenon that higher-accuracy models exhibit greater variance under imputation because they capture more spatially specific information. In contrast, the coarse 1~km SMOD layers exhibit artificially low variability because displaced clusters often fall within the same grid cell, producing near-identical evaluation outcomes across all imputations.

Beyond accuracy improvements, an important downstream application of the HUR dataset is its role as a high-resolution spatial prior. The map can support a wide range of empirical research by allowing users to infer the likely settlement type of a survey cluster, household or neighborhood. This is particularly relevant for studies in health, demography, poverty and migration, where stratification of sampling frames into rural and urban zones is central to the interpretation of outcomes. Because the HUR map is independent of DHS definitions and is spatially consistent across national boundaries, it provides a harmonized backdrop for comparative research.

From a policy and development perspective, the ability to classify non-human settlements, rural areas and urban zones with high spatial detail is crucial. Many African countries are experiencing rapid but highly heterogeneous settlement expansion, often characterized by informal peri-urban growth, ribbon development along transportation corridors, and the emergence of small service towns. These patterns frequently fall outside official administrative definitions of "urban," which historically rely on population thresholds or census-designated areas. The HUR maps offer a more spatially sensitive tool that can support monitoring of UN Sustainable Development Goals (SDGs)---especially SDG 11 (Sustainable Cities and Communities), SDG 3 (Good Health and Well-Being), and SDG 1 (No Poverty---and provides evidence for more targeted planning of infrastructure, emergency response systems, sanitation networks, and health service delivery.

A deliberate methodological choice in this study was to train the model using only binary rural-urban labels, corresponding directly to DHS definitions. This avoided the introduction of inconsistent or subjective classifications of peri-urban areas, which lack a unified global definition. While the binary structure provides a clear and conservative validation framework, it does not explicitly resolve finer settlement typologies such as peri-urban, semi-dense, or mixed-use areas. Developing a robust multi-class model remains an essential direction for future work and will require either newly curated in-situ labels or harmonized multi-source definitions that more explicitly capture rural-urban transitions.

Several future research directions emerge from the present findings. First, extending the temporal coverage beyond 2016–2022 would enable the study of long-term settlement evolution, including pre-2010 urbanization patterns observable through Landsat's historical archive. Second, incorporating additional sensing modalities, such as Sentinel-1 SAR for cloud-prone regions or socioeconomic proxies such as mobile-phone data, may improve classification reliability in difficult-to-model regions. Third, integrating HUR maps with survey-based indicators on health, wealth, and education could unlock new possibilities for spatially explicit development research and refine the understanding of spatial inequalities. Finally, extending this framework to other regions of the Global South would allow for global-scale harmonization and assessment of transferability across different landscape and settlement morphologies.

In summary, this study underscores the value of high-resolution, deep-learning-derived rural-urban maps for both scientific and policy applications. By overcoming the limitations of existing kilometer-scale products, the HUR maps provide a more accurate, spatially detailed and context-aware representation of Africa’s settlement landscape. This advancement is particularly relevant in settings experiencing rapid urbanization, evolving rural livelihoods, and increasing environmental pressures. As development challenges intensify and demand more precise spatial information, datasets such as HUR offer an essential foundation for monitoring change and informing equitable, data-driven policy decisions.

\subsection*{Data availability}
The dataset is freely available for download on the AI and Global Developmenet Lab website (global-lab.ai)  Figshare (\url{https://doi.org/10.7910/DVN/YFRECD}) \cite{DVN/YFRECD_2024}. It provides a high-resolution rural-urban map of the African continent, projected in the WGS84 coordinate system. The dataset covers the entire continent using multiple tiles, each with its respective Coordinate Reference System (CRS). It offers a spatial resolution of 10~m and is organized into 86 Geotiff tiles. The file naming convention follows the format: $RuralUrban\_aa\_COG.tif$, where $aa$ represents the tile number. Additionally, a PNG file named $AfricaGeometry.png$ is included to illustrate the tile layout across the continent.

We created an interactive GEE application  (\url{https://kakooeimohammd.users.earthengine.app/view/rural-urban-africa}) to facilitate data access. In this application, users can click on individual tiles to retrieve download links for the corresponding data. Each GeoTIFF file includes seven bands (b1 to b7), representing data from the years 2016 to 2022.

\subsection*{Code availability}
The results were produced using Python (v3.10.12) on NAISS and GEE. More information about the scripts are available on GitHub in the repository of the AI and Global Development Lab: \url{https://github.com/AIandGlobalDevelopmentLab/Africa-Rural-Urban-Map}.

\bibliography{refs,AdelReferences}

\section*{Acknowledgements}

This work was supported by the Swedish Research Council to the AI and Global Development Laboratory under Grant 2020-03088 and Grant 2020-00491. JB gratefully acknowledges partial financial support from the Australian-American Fulbright Commission and the Kinghorn Foundation.

The computations and data handling were enabled by resources provided by the NAISS, partially funded by the Swedish Research Council through grant agreement no. 2022-06725.

\section*{Author contributions statement}

Conceptualization: M.K. and A.D.; Methodology: A.S., A.B., J.B., M.P., and M.K.; Investigation: A.S., A.B., M.K., M.P., and A.D.; Writing—original draft: M.K.; Writing—review and editing: M.K., J.B., M.P., and A.D.; Supervision: M.K., and A.D.; Funding acquisition: A.D. All authors have read and agreed to the published version of the manuscript.

\section*{Additional information}

The authors declare no competing interests.

\end{document}